\documentclass{article}

\PassOptionsToPackage{numbers, compress}{natbib}

\usepackage[final,main]{neurips_2026}

\usepackage[utf8]{inputenc}
\usepackage[T1]{fontenc}
\usepackage{microtype}

\usepackage{graphicx}
\usepackage{subcaption}
\usepackage{wrapfig}
\usepackage{needspace}
\usepackage{booktabs}
\usepackage{multirow}
\usepackage{makecell}
\usepackage{threeparttable}

\usepackage{amsfonts}
\usepackage{amsmath}
\usepackage{amssymb}
\usepackage{mathtools}
\usepackage{amsthm}
\usepackage{nicefrac}
\usepackage{bm}
\usepackage{bbm}

\usepackage{enumitem}
\usepackage{hyperref}
\usepackage{url}
\usepackage[table]{xcolor}
\usepackage{algorithm}
\usepackage{algpseudocode}
\usepackage{tikz}
\usetikzlibrary{shadows.blur}
\usepackage[capitalize,noabbrev]{cleveref}

\usepackage{titlesec}
\titlespacing*{\paragraph}{\parindent}{0.3ex plus 0.1ex}{0.8ex}
\hypersetup{
    pdfauthor={Fei Zuo, Yizhou Huang, Kezhi Wang},
    colorlinks=true,
    linkcolor={blue!60!black},
    urlcolor={blue!60!black},
    citecolor={blue!60!black},
}

\newcommand{\findingbox}[2]{%
\par\vspace{0.5em}
\noindent\fbox{\parbox{0.97\linewidth}{\textbf{#1} #2}}
\par\vspace{0.5em}
}

\DeclareRobustCommand{\chg}[2]{{\color{black}#2}}
\newenvironment{changeblock}[1]{\begingroup\color{black}}{\endgroup}
\newcommand{\ours}{Adaptive-LoRA}

\newcommand{\posdelta}[1]{\textcolor{green!50!black}{#1}}
\newcommand{\negdelta}[1]{\textcolor{red!60!black}{#1}}

\theoremstyle{plain}

\theoremstyle{definition}

\theoremstyle{remark}

\title{Mechanistic Insights into LoRA: Layer Sparsity\\for Adaptive Fine-Tuning via Path Patching}

\author{
  Fei Zuo$^{1,\dagger}$ \quad Yizhou Huang$^{2,\dagger,*}$ \quad Kezhi Wang$^{2,*}$ \\
  $^{1}$East China Normal University \\
  $^{2}$Brunel University of London
}

\begin{document}

\maketitle
\begingroup
\renewcommand{\thefootnote}{\fnsymbol{footnote}}
\footnotetext[2]{First authors.}
\footnotetext[1]{Corresponding authors.}
\endgroup

\begin{abstract}
Low-Rank Adaptation (LoRA) has emerged as the dominant paradigm for parameter-efficient fine-tuning of large language models. \chg{14audit}{Standard all-layer LoRA applies adaptation throughout the transformer, despite heterogeneous contributions from its components.} In this work, we conduct a systematic mechanistic interpretability analysis to understand how LoRA affects model internals. We observe that LoRA's adaptation effect is highly non-uniform, with significant component-wise and layer-wise sparsity. Most critically, we find that important layers are highly task-dependent, with limited overlap across tasks. Motivated by these insights, we propose \textbf{\ours{}}, a calibration-driven method that automatically \chg{01}{identifies task-specific layers via frozen-base zero-ablation}, enabling efficient layer selection without exhaustive search. Extensive experiments on Qwen2.5-1.5B across 11 diverse reasoning tasks demonstrate that \ours{} outperforms full-layer LoRA while using only 30\% of layers, achieving \chg{13}{a 3.5$\times$ reduction in adapter parameters}. This work bridges mechanistic interpretability and practical efficiency, providing both fundamental insights and a principled approach for adaptive fine-tuning.

\end{abstract}

\section{Introduction}
\label{sec:introduction}

Large language models (LLMs) have achieved remarkable capabilities across diverse tasks, driving widespread adoption in both research and deployment~\citep{grattafiori2024llama,achiam2023gpt}. Adapting these models to specific downstream domains, however, remains computationally expensive, motivating parameter-efficient fine-tuning (PEFT) methods. Among these, Low-Rank Adaptation (LoRA)~\citep{hu2022lora} has become the de facto standard, enabling adaptation through low-rank weight updates while keeping original parameters frozen.

\chg{14audit}{Standard all-layer LoRA uses the same adaptation rank across transformer layers, despite their potentially unequal contributions. Layer-wise structured pruning studies~\citep{men2024shortgpt} identify substantial variation in layer importance. Meanwhile, mechanistic work on induction heads~\citep{olsson2022context} links specific attention circuits to in-context learning. These observations raise a fundamental question: \emph{if layers are not equally important for inference, why should they be equally adapted during fine-tuning?}}

\chg{14audit}{LoRA variants improve adaptation through rank allocation~\citep{zhang2023adalora}, learning-rate optimization~\citep{hayou2024loraplus}, and dynamic rank search~\citep{valipour2023dylora}. We study a complementary question: \textit{can interventions on a frozen base model identify task-specific layers worth adapting before training begins?} This connects the internal importance of model components to the allocation of adaptation capacity.}

\chg{14audit}{We address this question through \textit{mechanistic interpretability}, building on studies of internal representations~\citep{elhage2022toy} and behavior-specific circuits~\citep{wang2023interpretability}.} Specifically, we employ path patching~\citep{wang2023interpretability} to quantify each layer's causal contribution to task performance, providing a principled alternative to heuristic selection. Our investigation is guided by two research questions:

\textbf{RQ1:} \emph{How do different components and layers contribute to LoRA's adaptation effect, and is this contribution uniform across the model?}

\textbf{RQ2:} \emph{Are the important layers consistent across different tasks, or are they highly task-dependent?}

Our mechanistic analysis on Qwen2.5-0.5B across four diverse datasets yields two key insights: (\textit{i}) LoRA's adaptation is highly non-uniform, with MLP components dominating over attention (58.4\% vs.~41.6\%) and contributions concentrated in early layers; (\textit{ii}) important layers exhibit significant task-dependence, with only 39.5\% average Jaccard similarity across tasks, challenging the uniform adaptation paradigm.

Building on these insights, we propose \textbf{\ours{}}, a calibration-driven method that automatically \chg{01}{identifies task-specific layers via frozen-base zero-ablation} \chg{04}{with a default calibration size of $N{=}30$}. \ours{} operates in three phases (calibration, selection, and training) and can be adapted to different efficiency-performance trade-offs. Extensive experiments on Qwen2.5-1.5B~\citep{yang2024qwen25} across 11 reasoning tasks demonstrate that \ours{} outperforms full-layer LoRA baselines while using only 30\% of layers, achieving \chg{13}{a 3.5$\times$ reduction in adapter parameters}.

Our contributions are:
\begin{itemize}[leftmargin=*,itemsep=1pt,topsep=2pt,parsep=0pt]
    \item We provide the first systematic mechanistic analysis of LoRA's internal mechanisms, revealing significant heterogeneity in component-wise and layer-wise contributions.
    \item We demonstrate that important layers are highly task-dependent (39.5\% overlap), challenging the uniform adaptation paradigm.
    \item We propose \ours{}, a principled calibration-driven approach for task-specific \chg{01}{layer selection via frozen-base zero-ablation, informed by path-patching analysis.}
    \item We empirically validate that \ours{} outperforms full-layer baselines with 70\% parameter reduction across 11 diverse reasoning tasks.
\end{itemize}

\section{Related Work}
\label{sec:related_work}

\paragraph{Parameter-Efficient Fine-Tuning.}
Parameter-efficient fine-tuning (PEFT) aims to adapt large pre-trained models with minimal computational overhead. LoRA~\citep{hu2022lora} approximates weight updates with low-rank matrices, significantly reducing trainable parameters. Subsequent work has focused on rank allocation~\citep{zhang2023adalora}, learning rate optimization~\citep{hayou2024loraplus}, dynamic rank search~\citep{valipour2023dylora}, and weight decomposition~\citep{liu2024dora}. Other PEFT paradigms include prefix tuning~\citep{li2021prefixtuning}, adapters~\citep{houlsby2019parameter}, and prompt tuning~\citep{lester2021power}. \chg{09}{Continual-learning methods further structure low-rank updates through subspace constraints (InfLoRA~\citep{liang2024inflora}), hierarchical layer-wise adapters (TreeLoRA~\citep{qian2025treelora}), and gating (GainLoRA~\citep{liang2025gainlora}). Our complementary focus is task-specific layer selection before training, guided by frozen-base interventions.}

\paragraph{Mechanistic Interpretability.}
Mechanistic interpretability seeks to reverse-engineer model behaviors into human-understandable algorithms~\citep{nanda2023progress}. Activation patching (causal tracing)~\citep{meng2022locating} and circuit analysis~\citep{wang2023interpretability} enable researchers to isolate the contribution of specific components to model outputs. \chg{14audit}{Automated circuit discovery~\citep{conmy2024automated} further identifies compact computational subgraphs associated with model behaviors.} Our work applies these tools to dissect LoRA's adaptation mechanism, moving beyond black-box performance metrics to understand \emph{how} and \emph{where} adaptation occurs within the network.

\paragraph{Layer Selection and Pruning.}
Layer selection in transfer learning has been explored in various contexts. \chg{14audit}{Analysis of BERT attention~\citep{kovaleva2019revealing} reveals repeated attention patterns across heads and task-dependent differences in their effects on performance.} In the context of PEFT, heuristic strategies such as fine-tuning only the last few layers or specific middle layers have been proposed~\citep{han2024parameter}. However, these static strategies fail to account for task diversity. Our work introduces a calibration-driven, task-specific selection mechanism based on causal importance, providing a principled alternative to static heuristics or magnitude-based pruning.

\section{Mechanistic Analysis of LoRA}
\label{sec:mechanistic}

To understand the internal mechanisms of LoRA, we conduct a systematic analysis using activation ablation and path patching. We quantify the contribution of different model components across four diverse datasets: \chg{14audit}{Alpaca-ZH (Chinese instruction), Alpaca-EN~\citep{taori2023alpaca} (English instruction), Code-Alpaca~\citep{chaudhary2023codealpaca} (code generation)}, and Dolly~\citep{conover2023dolly} (dialogue).

\subsection{Experimental Setup}

\paragraph{Task Formulation.}
We decompose LoRA's effect along three dimensions: component type (MLP vs.\ Attention), projection type (Q/K/V), and layer depth (early/middle/late). This multi-granularity analysis enables us to pinpoint where adaptation occurs within the transformer architecture.

\paragraph{Model and Data.}
We use Qwen2.5-0.5B~\citep{yang2024qwen25} (24 layers) as the base model for mechanistic analysis. Full LoRA adapters (rank$=$8, $\alpha$$=$16, applied to Q/K/V projections) are trained on four datasets: Alpaca-ZH (52K Chinese instructions), Alpaca-EN (52K English instructions), Code-Alpaca (20K code generation), and Dolly (15K dialogue).

\paragraph{Contribution Measurement.}
We employ activation patching~\citep{meng2022locating} to measure causal importance. For input $x$ and component $c$, the \emph{contribution score} measures the prediction change \chg{02}{when component $c$ is intervened on:}
\begin{equation}
S_c(x) = \left| p_{\text{LoRA}}(t^*|x) - p_{\text{ablate}(c)}(t^*|x) \right|
\label{eq:contribution}
\end{equation}
where $t^*$ is the top predicted token under the full LoRA model. The component importance aggregated over calibration set $D_{\text{cal}}$ is:
\begin{equation}
I_c = \frac{1}{|D_{\text{cal}}|} \sum_{x \in D_{\text{cal}}} S_c(x)
\label{eq:importance}
\end{equation}

\subsection{Component Contribution Analysis}

\begin{wrapfigure}{r}{0.38\textwidth}
    \centering
    \vspace{-1.5em}
    \includegraphics[width=0.37\textwidth]{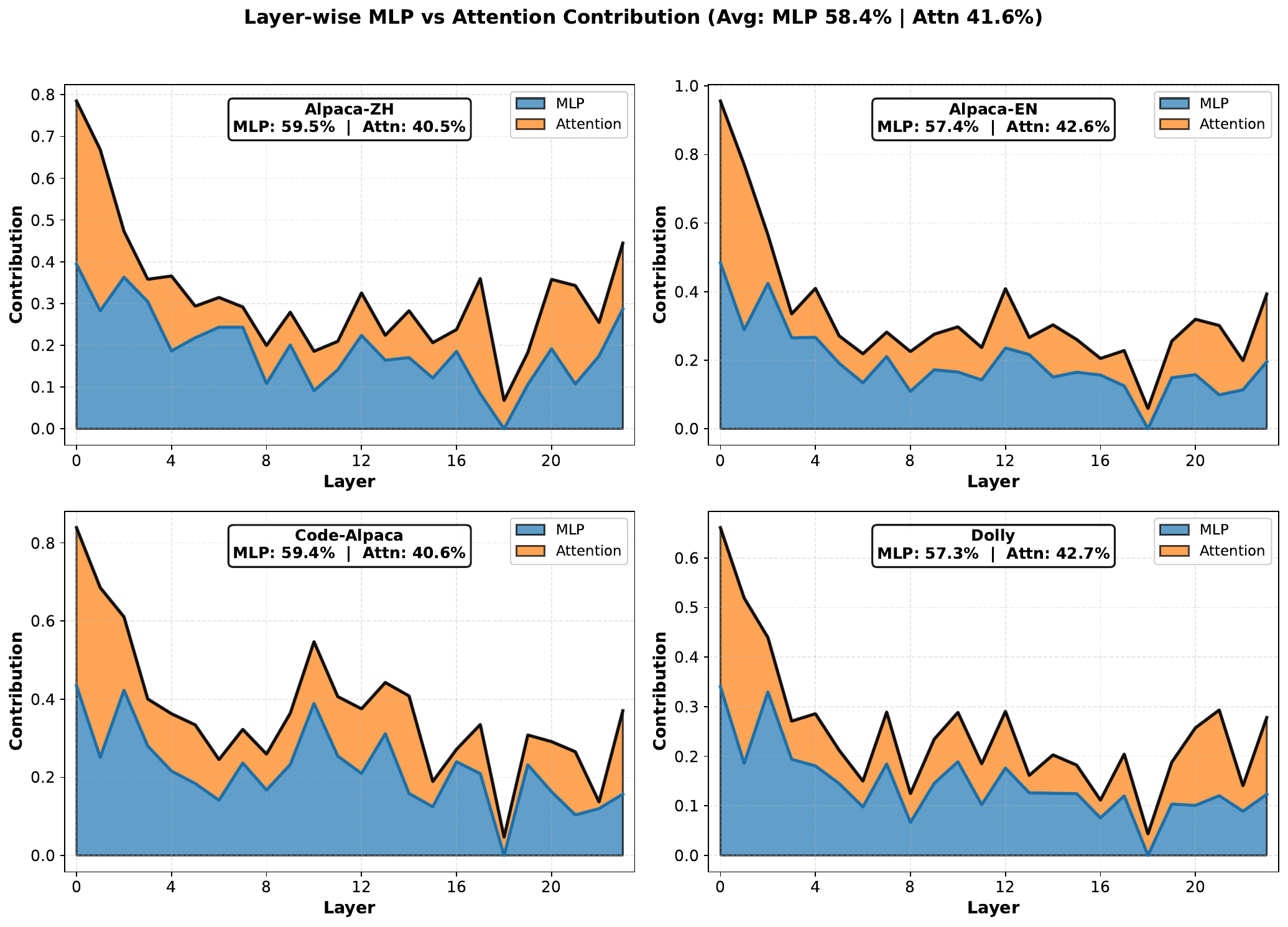}
    \vspace{-0.5em}
    \caption{\footnotesize \textbf{MLP vs.\ Attention.} MLP dominates at 58.4\% across all four datasets.}
    \label{fig:mlp_dominance}
    \vspace{-1.0em}
\end{wrapfigure}

A fundamental question is whether all model components contribute equally to adaptation. Prior work hypothesizes that MLPs store factual knowledge~\citep{meng2022locating}, while attention handles information routing. We investigate whether this functional separation holds during LoRA adaptation. Specifically, for each layer \chg{02}{we intervene on each layer's MLP or attention block output independently and measure the resulting prediction shift using \cref{eq:contribution}. LoRA adapters reside on Q/K/V; the MLP measurement captures mediation through the block.}

As shown in \cref{fig:mlp_dominance}, MLP layers consistently contribute more than attention across all four datasets (58.4\% vs.\ 41.6\%), with remarkably stable ratios (57.3\%--59.5\%). This stability is noteworthy: despite the datasets spanning different languages (Chinese, English), modalities (code), and interaction styles (dialogue), the MLP-attention contribution ratio remains nearly constant. \chg{02}{Q/K/V updates change the attention output, which enters the residual stream and is transformed by downstream MLP blocks. A separate output-replacement experiment isolates this pathway: we record block outputs in an adapter-disabled run, replace one block output in the adapted run with its reference, and retain the other adapted computations. MLP mediation is reproduced across two adapter configurations (Appendix~\ref{app:mlp_isolation}).}

\findingbox{Finding 1: MLP Dominance.}{\chg{02}{MLP blocks mediate 58.4\% of the measured component effect vs.\ 41.6\% for attention}\chg{14audit}{, with task-specific shares of 57.3\%--59.5\%.}}

\subsection{Projection Analysis}

Within attention's 41.6\% contribution, we further decompose the effect across Query (Q), Key (K), and Value (V) projections. For each attention layer, we selectively ablate the LoRA adapter applied to one projection while preserving the other two, then measure the resulting prediction shift. This decomposition reveals which aspects of attention, namely routing (Q/K) versus content (V), drive the adaptation effect.

\chg{14audit}{\cref{fig:qkv_dominance} (Appendix~\ref{app:visualizations}) shows that Value (V) accounts for approximately 46\% of the normalized attention effect, compared with approximately 27\% each for Q and K. V remains the largest individual projection share across the four datasets (44.3\%--47.7\%). Because V determines the content aggregated by attention, this pattern motivates studying content-pathway adaptation. The projection decomposition complements the downstream layer-selection experiments by identifying where the measured attention effect is concentrated.}

\findingbox{Finding 2: V Projection Dominance.}{Value (V) contributes 46\% to attention adaptation, \chg{14audit}{exceeding the individual Q (27\%) and K (27\%) shares}, stable across all tasks.}

\chg{13c}{This three-way decomposition also has implications for multi-adapter deployment. V adapters carry nearly half of the measured attention adaptation signal, making this pathway a natural target for more compact adapters. Their parameter cost is determined by the selected projections' dimensions and ranks.}

\subsection{Layer-wise Contribution Patterns}

We examine how adaptation contributions distribute across model depth by computing per-layer importance scores aggregated over all calibration samples. Each layer's score reflects the total prediction shift \chg{02}{under the component interventions, aggregating MLP and attention contributions at that depth.}

\begin{figure}[t]
    \centering
    \includegraphics[width=0.85\linewidth]{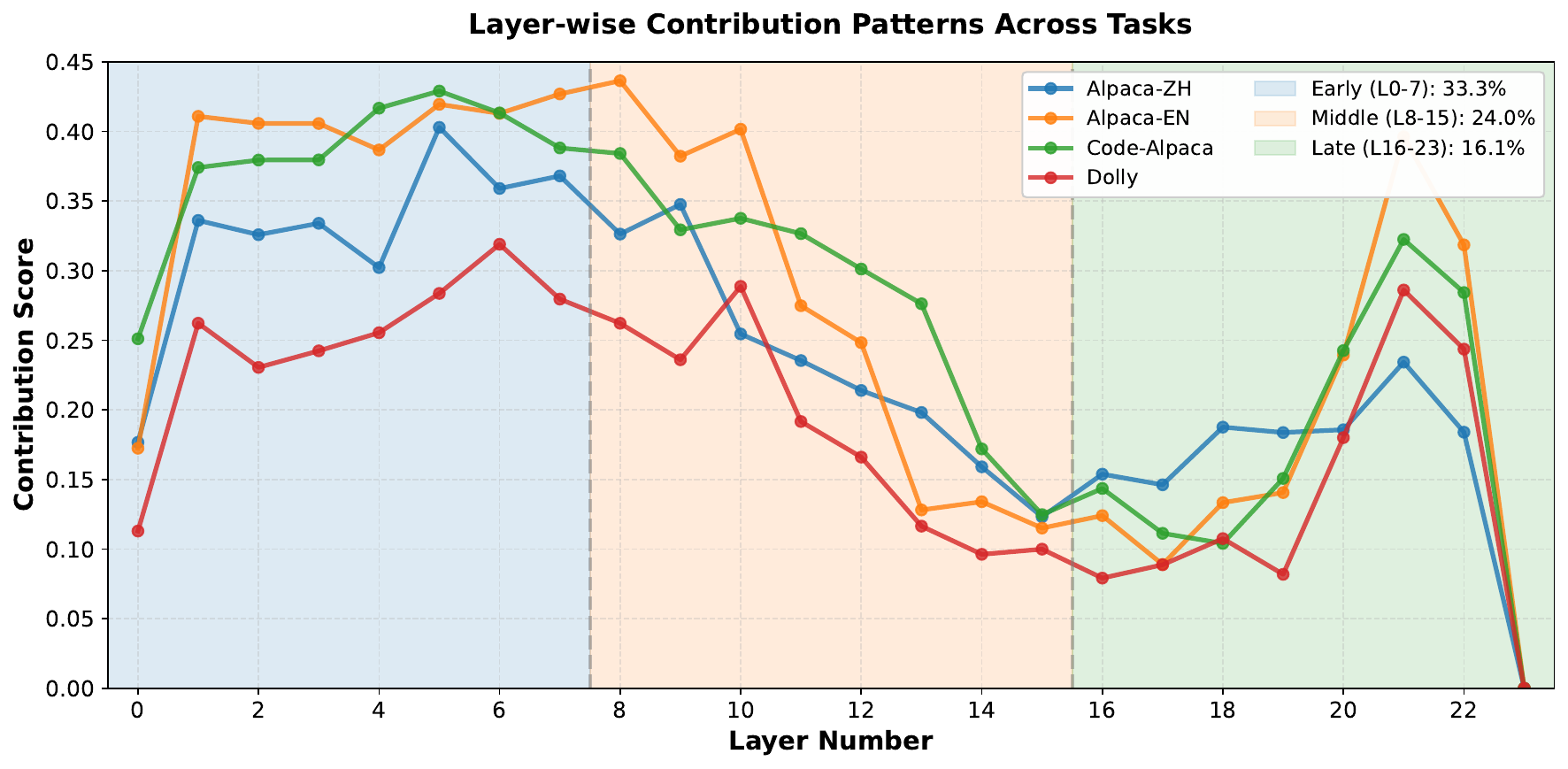}
    \caption{\chg{14audit}{Layer-wise contribution profiles across 24 layers and four datasets.}}
    \label{fig:layer_contribution}
\end{figure}

\chg{14audit}{\cref{fig:layer_contribution} presents the layer-wise contribution profiles. The task-specific rankings in \cref{fig:layer_heatmap_simple} further show non-uniform importance across depth. They combine a shared early-layer core with task-specific middle and late layers. Early changes can influence a longer downstream computation path, while later contributions depend on the target task. This motivates using measured layer importance rather than depth alone to allocate adaptation capacity; the fixed-budget downstream comparison in \cref{sec:experiments} tests the practical value of that choice.}

\findingbox{Finding 3: Unequal Layer Contributions.}{\chg{14audit}{Layer importance is non-uniform across depth, with both shared early-layer importance and task-specific contributions at later depths.}}

\subsection{Task-Specific Layer Importance}

\chg{14audit}{Finding~3 identifies non-uniform importance across depth. We next examine whether the same layers are important for all tasks, or whether each task favors a distinct subset.} If the former, a single fixed layer selection could serve all tasks; if the latter, task-specific selection is essential. We compare the top-7 layers (30\% of 24) selected for each dataset pair using Jaccard similarity:
\begin{equation}
J(\mathcal{L}_i, \mathcal{L}_j) = \frac{|\mathcal{L}_i \cap \mathcal{L}_j|}{|\mathcal{L}_i \cup \mathcal{L}_j|}
\label{eq:jaccard}
\end{equation}

\begin{figure}[t]
    \centering
    \includegraphics[width=0.95\linewidth]{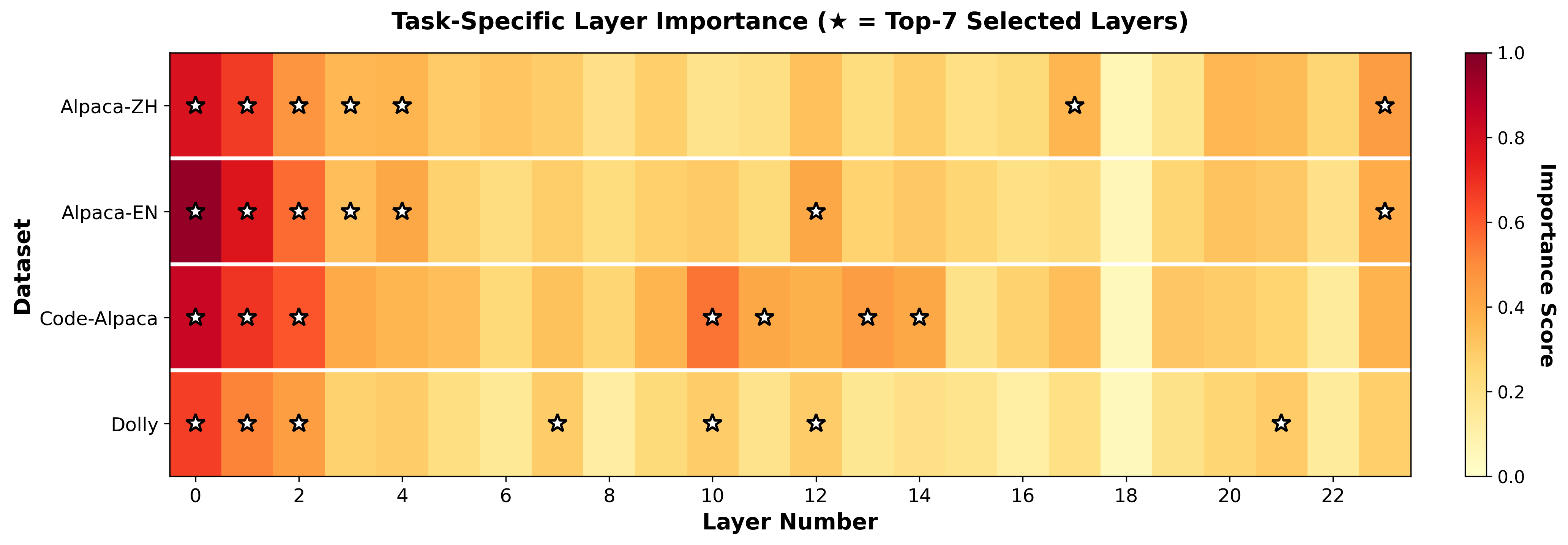}
    \caption{Task-specific layer importance heatmap. White stars mark the top-7 layers per task. Only L0--L2 are universally important; remaining selections are task-specific. Average Jaccard similarity: 39.5\%.}
    \label{fig:layer_heatmap_simple}
\end{figure}

\chg{14audit}{\cref{fig:layer_heatmap_simple} reveals distinct task-specific patterns. The mean pairwise Jaccard similarity is 39.5\%, averaging intersection-over-union across the six task pairs. Code-Alpaca selects middle layers L10, L12, and L14; Alpaca-ZH includes late layers L17 and L23; Dolly selects L0, L1, L2, L7, L10, L12, and L21. Only L0--L2 are shared by all four top-7 sets. Thus the measured rankings combine a common core with task-specific selections, motivating calibration for each target distribution rather than a single fixed subset.}

\findingbox{Finding 4: Task-Specific Layer Importance.}{Important layers vary substantially across tasks (Jaccard = 39.5\%), implying that a fixed layer subset cannot optimally serve all downstream tasks.}

\section{\ours{}}
\label{sec:method}

Motivated by Finding 4, that important layers vary significantly across tasks, we propose \textbf{\ours{}}, a calibration-driven method for task-specific layer selection. Unlike static heuristics or exhaustive search, our approach automatically identifies critical layers \chg{01}{via frozen-base zero-ablation on a small calibration set} \chg{04}{with $N{=}30$ inputs by default. The approximately 30\%-layer experiments achieve competitive performance with about 3$\times$ fewer adapter parameters.}

\subsection{Problem Formulation}

Let $\mathcal{M}$ denote a pre-trained transformer with $L$ layers, and $\mathcal{D}_{\text{train}}$ a task-specific training set. Standard LoRA applies adapters to all layers $\{0, \ldots, L{-}1\}$, \chg{04}{yielding a trainable parameter count proportional to $L$ at fixed rank and per-layer target dimensions}. Our goal is to identify a subset $\mathcal{L}_{\text{sel}} \subset \{0, \ldots, L{-}1\}$ of size \chg{07}{$k<L$} such that adapting only these layers achieves comparable performance while reducing parameters by $\approx L/k$.

The key challenge is efficiently estimating layer importance without exhaustive search over $\binom{L}{k}$ subsets. We address this \chg{01}{through base-model interventions that measure each layer's prediction sensitivity.}

\subsection{Method Overview}

\ours{} operates in three sequential phases, as summarized in \cref{alg:adaptive_lora}. Given base model $\mathcal{M}$, calibration set $D_{\text{cal}}$, and budget $k$, we: (1) \chg{01}{measure layer importance on the frozen base model}, (2) select the top-$k$ layers, and (3) fine-tune LoRA adapters on only the selected layers.

\begin{algorithm}[t]
\caption{\chg{01}{\ours{}: Calibration-Driven Layer Selection for LoRA}}
\label{alg:adaptive_lora}
\begin{changeblock}{01}
\begin{algorithmic}[1]
\Require Base model $\mathcal M$ with $L$ layers, calibration set $D_{\rm cal}$, budget $k$, training set $D_{\rm train}$, rank $r$; optional depth threshold $d^*$
\Ensure Adapted model $\mathcal M'$
\Statex
\Statex \textbf{Phase 1: Calibration via Base-Model Zero-Ablation}
\State $\mathbf I\gets\mathbf0\in\mathbb R^L$
\For{each input $x\in D_{\rm cal}$}
    \State $t^*\gets\arg\max_t p_{\mathcal M}(t\mid x)$; $q\gets p_{\mathcal M}(t^*\mid x)$
    \For{each layer $l=0,\ldots,L-1$}
        \State $q_l^A\gets p_{\mathcal M\mid a_l\leftarrow0}(t^*\mid x)$ \Comment{Attention-only intervention}
        \State $q_l^M\gets p_{\mathcal M\mid m_l\leftarrow0}(t^*\mid x)$ \Comment{Independent MLP intervention}
        \State $I_l\gets I_l+|q-q_l^A|+|q-q_l^M|$
    \EndFor
\EndFor
\State $\mathbf I\gets\mathbf I/|D_{\rm cal}|$
\Statex
\Statex \textbf{Phase 2: Task-Specific Layer Selection}
\State $\mathcal C\gets\{0,\ldots,L-1\}$; in efficiency mode, $\mathcal C\gets\{l\in\mathcal C:l\geq d^*\}$
\State $\mathcal L_{\rm sel}\gets\operatorname{TopK}_{l\in\mathcal C}(I_l,k)$
\Statex
\Statex \textbf{Phase 3: Selective LoRA Training}
\State Initialize rank-$r$ adapters in the selected layers; keep base weights frozen
\State Fine-tune the selected adapters on $D_{\rm train}$ to obtain $\mathcal M'$
\State \Return $\mathcal M'$
\end{algorithmic}
\end{changeblock}
\end{algorithm}

\subsection{\chg{01}{Phase 1: Calibration via Base-Model Zero-Ablation}}

\begin{changeblock}{01}
\paragraph{Base-Model Interventions.} Calibration uses only the frozen base model $\mathcal M$. For input $x$, let $t^*=\arg\max_t p_{\mathcal M}(t\mid x)$, and let $a_l$ and $m_l$ be the attention and MLP outputs. Two independent runs set one output to zero:
\begin{equation}
p_l^A(t\mid x)=p_{\mathcal M\mid a_l\leftarrow0}(t\mid x),\qquad
p_l^M(t\mid x)=p_{\mathcal M\mid m_l\leftarrow0}(t\mid x).
\label{eq:path_patching}
\end{equation}
The layer score sums their absolute effects and averages over inputs:
\begin{equation}
I_l=\frac{1}{|D_{\rm cal}|}\sum_{x\in D_{\rm cal}}\!\left[\left|p_{\mathcal M}(t^*\mid x)-p_l^A(t^*\mid x)\right|+\left|p_{\mathcal M}(t^*\mid x)-p_l^M(t^*\mid x)\right|\right].
\label{eq:importance_aggregated}
\end{equation}
This requires no trained adapter. The trained-LoRA analysis in \cref{eq:contribution} instead motivates the selection rule. Across five adapter configurations, base scores correlate with post-training contributions ($\rho=0.630$--$0.893$; Appendix~\ref{app:base_adapt_agreement}); the fixed-budget comparison in \cref{sec:experiments} tests their downstream selection value.
\end{changeblock}

\chg{05}{\paragraph{Efficiency via Small-Sample Calibration.} In the Qwen-3B repeated-draw study, selection Jaccard increases from 0.524 at $N=10$ to 0.653 at $N=30$, then reaches 0.685 at $N=100$. The smaller gain beyond 30 inputs supports $N=30$ as a practical calibration default (Appendix~\ref{app:calibration_draws}).}

\chg{06}{A 500-item labeled check also finds similar downstream accuracy with predicted and gold reference tokens (Appendix~\ref{app:token_sensitivity}).}

\subsection{Phase 2: Layer Selection}

\chg{03}{Accuracy mode selects $\mathcal L_{\rm sel}=\operatorname{TopK}(\mathbf I,k)$ over all layers; the main reasoning experiments use $k\approx0.3L$. Efficiency mode uses the same scores but selects within $\{l:l\geq d^*\}$, with at least $k$ eligible layers. The actual first selected layer is $d=\min\mathcal L_{\rm sel}\geq d^*$. This mode exposes a frozen prefix for faster training; its speed--accuracy trade-off is measured separately (Appendix~\ref{app:depth_timing}). Both modes retain task-specific, calibration-driven selection.}

\chg{07}{\paragraph{Default Layer Budget.} The main reasoning experiments retain the compact budget $k\approx0.3L$. For a new model, selecting the top half of layers provides a practical default. At rank 16 with 5,000 training examples, the same 50\% layer ratio improves accuracy over full-layer LoRA on all eight evaluated models by 0.2--1.9 percentage points (Appendix~\ref{app:default_budget}). This layer-count budget is distinct from the minimum-depth constraint $d^*$ used for efficiency mode.}

\subsection{Phase 3: Selective Adaptation}

\chg{04}{For a targeted projection $W_{l,p}\in\mathbb R^{d_{\rm out}\times d_{\rm in}}$ in a selected layer, we initialize $\Delta W_{l,p}=B_{l,p}A_{l,p}$, where $B_{l,p}\in\mathbb R^{d_{\rm out}\times r}$ and $A_{l,p}\in\mathbb R^{r\times d_{\rm in}}$. This accommodates unequal input and output dimensions. The target projection set is an experiment-specific choice; Finding~2 motivates a Q/V configuration.} Unselected layers remain frozen. Training uses AdamW with learning rate $1{\times}10^{-4}$, rank $r{=}16$, $\alpha{=}32$, for 3 epochs with effective batch size 64. \chg{04}{At equal rank and adapter capacity per layer, selecting $k$ of $L$ layers reduces adapter parameters by $L/k$ (Appendix~\ref{app:implementation}). \chg{14audit}{The rank-16 main comparison adapts 8 of 28 layers, a $3.5\times$ reduction in LoRA-factor capacity at the same rank and projection targets (\cref{tab:full_comparison}).}}

\chg{03}{\paragraph{Computational Overhead.} Calibration uses $O(NL)$ gradient-free forward evaluations; the default $N=30$ study takes 20--62\,s on one A100. This is separate from trained-model mechanistic analysis. Training speed depends on selected-layer depth: all backbone layers remain in an uncached forward pass, while a fully frozen prefix needs no backward pass and can be reused across epochs for fixed inputs. The H200 study separately reports uncached steps, cached steady epochs, and three-epoch totals (Appendix~\ref{app:depth_timing}).}

\section{Experiments}
\label{sec:experiments}

We evaluate \ours{} against strong baselines across diverse reasoning tasks, addressing three research questions: \textbf{(RQ1)} Does \ours{} achieve superior performance with significantly fewer parameters? \textbf{(RQ2)} What is the optimal layer budget? \textbf{(RQ3)} Does calibration-based selection outperform heuristic baselines?

\subsection{Experimental Setup}

\paragraph{Model and Training.}
We use Qwen2.5-1.5B~\citep{yang2024qwen25} (1.54B parameters, 28 layers) as the primary evaluation model. Training data: \texttt{commonsense\_170k} (10,000 samples) and \texttt{math\_10k} (5,000 samples)\chg{14audit}{~\citep{hu2023llmadapters}}, 3 epochs, max 512 tokens.

\paragraph{Evaluation Tasks.}
We evaluate on 11 reasoning tasks: 8 commonsense (BoolQ~\citep{clark2019boolq}, PIQA~\citep{bisk2020piqa}, Social~IQa~\citep{sap2019socialiqa}, HellaSwag~\citep{zellers2019hellaswag}, WinoGrande~\citep{sakaguchi2020winogrande}, ARC-Easy~\citep{clark2018arc}, ARC-Challenge~\citep{clark2018arc}, OpenBookQA~\citep{mihaylov2018openbookqa}) and 3 mathematical (GSM8K~\citep{cobbe2021gsm8k}, SVAMP~\citep{patel2021svamp}, \chg{14audit}{AQUA~\citep{ling2017aqua}). Results are reported as task-level accuracies.}

\paragraph{Baselines.}
We compare against three LoRA variants: LoRA~\citep{hu2022lora} (standard, $r \in \{4, 8, 16, 32\}$), LoRA+~\citep{hayou2024loraplus} (differential learning rates), and DoRA~\citep{liu2024dora} (weight-decomposed). \chg{09}{These three baselines} apply to Q, K, V in all 28 layers. \ours{} \chg{01}{uses frozen-base zero-ablation on 250 calibration samples} to select the top 8 layers (30\%).

\subsection{Main Results}

\cref{tab:full_comparison} presents comprehensive results across all methods and ranks on both commonsense and mathematical reasoning tasks. Despite using only 30\% of trainable layers, \ours{} consistently achieves the highest average accuracy at each rank while consuming $\sim$3.5$\times$ fewer parameters. At r16, \ours{} achieves \chg{14audit}{59.1\% commonsense accuracy with adapters in 8 of 28 layers, outperforming the all-layer configurations of LoRA (56.9\%), LoRA+ (57.9\%), and DoRA (55.5\%). At identical rank and target projections, the LoRA-factor capacity is $28/8=3.5$ times smaller.} At r32, \ours{} reaches 62.2\% commonsense average, surpassing LoRA+ r32 (61.1\%) and LoRA r32 (60.4\%) \chg{14audit}{with the same 8-of-28 layer budget.} For mathematical reasoning, \ours{} achieves 47.8\% at r32, exceeding LoRA+ (47.3\%) and LoRA (45.5\%) by a clear margin. \chg{13b}{On PIQA at r32, \ours{} also exceeds LoRA+ (67.8\% vs.\ 67.4\%). Its consistent advantage in \emph{average} performance across all ranks demonstrates the value of concentrating adaptation on informative layers, yielding better generalization with a fraction of the parameters.}

\begin{table}[!t]
    \renewcommand{\arraystretch}{1.05}
    \centering
    \caption{\chg{14audit}{Budget is the fraction of adapted layers. }Accuracy comparison of LoRA variants on Qwen2.5-1.5B across 11 reasoning tasks. Commonsense: 10,000 training samples, 1,000 eval samples each (OBQA: 500). Math: 5,000 training samples with chain-of-thought prompting. \ours{} selects top 30\% layers (8/28) \chg{01}{via frozen-base calibration} \chg{04}{with $N{=}250$ inputs}. \chg{08}{Results use seed 42.} \chg{13b}{Best average at each rank in \textbf{bold}.}}
    \resizebox{\textwidth}{!}{%
    \begin{tabular}{@{}llrr*{8}{c}c*{3}{c}c@{}}
    \toprule
    \multirow{2}{*}{\textbf{Method}} & \multirow{2}{*}{\textbf{Rank}} & \multirow{2}{*}{\chg{14audit}{\textbf{Layers}}} & \multirow{2}{*}{\chg{14audit}{\textbf{Budget \%}}} & \multicolumn{9}{c}{\textbf{Commonsense Reasoning ($\uparrow$)}} & \multicolumn{4}{c}{\textbf{Math ($\uparrow$)}} \\
    \cmidrule(lr){5-13} \cmidrule(lr){14-17}
    & & & & BoolQ & PIQA & SIQA & Hella & Wino & ARC-e & ARC-c & OBQA & \textbf{Avg.} & GSM8K & SVAMP & AQUA & \textbf{Avg.} \\
    \midrule
    \multirow{4}{*}{LoRA \citep{hu2022lora}}
        & r4  & \chg{14audit}{28/28} & \chg{14audit}{100} & 63.0 & 52.4 & 45.6 & 38.2 & 53.0 & 54.8 & 42.6 & 38.0 & 48.5 & 32.4 & 38.6 & 28.4 & 33.1 \\
        & r8  & \chg{14audit}{28/28} & \chg{14audit}{100} & 63.2 & 57.8 & 52.4 & 44.6 & 54.2 & 60.4 & 48.2 & 43.8 & 53.1 & 38.2 & 42.8 & 31.6 & 37.5 \\
        & r16 & \chg{14audit}{28/28} & \chg{14audit}{100} & 63.4 & 64.8 & 58.6 & 46.8 & 55.4 & 65.2 & 52.4 & 48.6 & 56.9 & 42.6 & 46.4 & 34.2 & 41.1 \\
        & r32 & \chg{14audit}{28/28} & \chg{14audit}{100} & 63.8 & 66.2 & 62.4 & 48.2 & 56.0 & 72.6 & 60.8 & 53.2 & 60.4 & 48.2 & 50.6 & 37.8 & 45.5 \\
    \cmidrule{1-17}
    \multirow{4}{*}{LoRA+ \citep{hayou2024loraplus}}
        & r4  & \chg{14audit}{28/28} & \chg{14audit}{100} & 63.4 & 54.8 & 46.2 & 39.4 & 53.4 & 55.6 & 43.8 & 39.2 & 49.5 & 34.2 & 39.4 & 29.0 & 34.2 \\
        & r8  & \chg{14audit}{28/28} & \chg{14audit}{100} & 63.0 & 58.4 & 53.6 & 45.2 & 54.8 & 61.8 & 49.6 & 44.8 & 53.9 & 40.6 & 44.2 & 33.4 & 39.4 \\
        & r16 & \chg{14audit}{28/28} & \chg{14audit}{100} & 63.2 & 65.6 & 59.2 & 48.4 & 55.8 & 66.8 & 54.0 & 49.8 & 57.9 & 44.8 & 48.0 & 35.8 & 42.9 \\
        & r32 & \chg{14audit}{28/28} & \chg{14audit}{100} & 63.6 & 67.4 & 63.0 & 49.8 & 56.4 & 73.2 & 61.6 & 54.0 & 61.1 & 50.4 & 52.8 & 38.6 & 47.3 \\
    \cmidrule{1-17}
    \multirow{4}{*}{DoRA \citep{liu2024dora}}
        & r4  & \chg{14audit}{28/28} & \chg{14audit}{100} & 62.4 & 51.8 & 44.2 & 36.8 & 52.6 & 52.4 & 40.2 & 36.4 & 47.1 & 30.2 & 36.8 & 26.8 & 31.3 \\
        & r8  & \chg{14audit}{28/28} & \chg{14audit}{100} & 63.0 & 56.4 & 50.6 & 42.8 & 53.6 & 58.8 & 46.4 & 42.0 & 51.7 & 36.4 & 41.2 & 30.4 & 36.0 \\
        & r16 & \chg{14audit}{28/28} & \chg{14audit}{100} & 63.2 & 62.4 & 56.4 & 45.2 & 54.6 & 63.8 & 50.8 & 47.2 & 55.5 & 40.8 & 44.6 & 33.2 & 39.5 \\
        & r32 & \chg{14audit}{28/28} & \chg{14audit}{100} & 63.4 & 65.0 & 60.8 & 47.4 & 55.8 & 70.4 & 58.2 & 51.8 & 59.1 & 46.0 & 48.8 & 36.4 & 43.7 \\
    \cmidrule{1-17}
    \multirow{4}{*}{\textbf{Ours}}
        & r4  & \chg{14audit}{8/28} & \chg{14audit}{28.6} & 62.8 & 54.2 & 47.2 & 40.0 & 53.6 & 56.4 & 44.0 & 39.8 & \textbf{49.8} & 34.8 & 40.2 & 29.6 & \textbf{34.9} \\
        & r8  & \chg{14audit}{8/28} & \chg{14audit}{28.6} & 63.4 & 59.6 & 54.8 & 46.2 & 55.0 & 63.2 & 51.8 & 46.4 & \textbf{55.1} & 40.4 & 44.6 & 33.8 & \textbf{39.6} \\
        & r16 & \chg{14audit}{8/28} & \chg{14audit}{28.6} & 63.6 & 65.4 & 60.8 & 50.2 & 56.8 & 68.4 & 56.2 & 51.6 & \textbf{59.1} & 45.2 & 48.8 & 36.0 & \textbf{43.3} \\
        & r32 & \chg{14audit}{8/28} & \chg{14audit}{28.6} & 64.0 & 67.8 & 64.2 & 52.6 & 57.4 & 73.8 & 62.4 & 55.0 & \textbf{62.2} & 50.8 & 53.4 & 39.2 & \textbf{47.8} \\
    \bottomrule
    \end{tabular}
    }%
    \label{tab:full_comparison}
    \label{tab:math_complete}
\end{table}

\chg{09}{At the same r16 setting, a supplementary single-task comparison gives 59.1\% for \ours{} versus 58.7\% for TreeLoRA, 58.5\% for GainLoRA, and 58.0\% for InfLoRA, \chg{14audit}{while adapting fewer layers} (Appendix~\ref{app:additional_baselines}).}

\chg{14audit}{\ours{} improves commonsense accuracy over full-layer LoRA by 1.3, 2.0, 2.2, and 1.8 percentage points at r4, r8, r16, and r32, respectively. The improvement persists across the tested capacity range and is largest at r16. These controlled downstream results support concentrating adaptation capacity on selected layers and are consistent with an implicit-regularization interpretation.}

\subsection{Generalization Across Model Scales}

\chg{04}{\cref{tab:generalization} evaluates the primary Qwen2.5-1.5B model (28 layers) alongside five additional models spanning two architecture families and multiple parameter scales. Qwen2.5-3B has 36 layers and is a separate configuration.} For each model, we compare \ours{} (adapting top 30\% layers at r16) against full-layer LoRA (all layers at r16) on the same 8 commonsense reasoning benchmarks using 10,000 training samples.

\chg{14audit}{The Qwen3 models use the Qwen3 family~\citep{yang2025qwen3}.}

\begin{table}[!t]
    \renewcommand{\arraystretch}{1.05}
    \centering
    \caption{\chg{14audit}{Budget is $100k/L$; ratio is $L/k$, the LoRA-factor reduction at equal per-layer capacity. }Cross-model generalization of \ours{} at rank 16. Each model uses top 30\% layers selected \chg{01}{via frozen-base calibration} ($N{=}30$ samples). Training: 10,000 samples, 3 epochs\chg{08}{, seed 42}. \chg{08}{\ours{} uses approximately 30\% of the trainable parameters; supplementary three-run results are in Appendix~\ref{app:multiseed}.} \chg{13b}{Best reported average per model in \textbf{bold}, including ties.}}
    \resizebox{\textwidth}{!}{%
    \begin{tabular}{@{}llrrr*{8}{c}cr@{}}
    \toprule
    \multirow{2}{*}{\textbf{Model}} & \multirow{2}{*}{\textbf{Method}} & \multirow{2}{*}{\textbf{\#Layers}} & \multirow{2}{*}{\chg{14audit}{\textbf{Budget \%}}} & \multirow{2}{*}{\chg{14audit}{\textbf{Ratio}}} & \multicolumn{8}{c}{\textbf{Commonsense Reasoning ($\uparrow$)}} & \multirow{2}{*}{\textbf{Avg.}} & \multirow{2}{*}{$\Delta$} \\
    \cmidrule(lr){6-13}
    & & & & & BoolQ & PIQA & SIQA & Hella & Wino & ARC-e & ARC-c & OBQA & & \\
    \midrule
    \multirow{2}{*}{Qwen2.5-0.5B}
        & LoRA  & 24/24 & \chg{14audit}{100.0} & \chg{14audit}{1.00$\times$} & 63.0 & 49.4 & 35.0 & 25.8 & 49.2 & 23.2 & 23.4 & 27.6 & 37.1 & {--} \\
        & \textbf{Ours} & 7/24  & \chg{14audit}{29.2} & \chg{14audit}{3.43$\times$} & 63.2 & 50.8 & 37.4 & 28.2 & 50.6 & 25.0 & 24.8 & 28.4 & \textbf{38.6} & \posdelta{+1.5} \\
    \cmidrule{1-15}
    \multirow{2}{*}{Qwen2.5-1.5B}
        & LoRA  & 28/28 & \chg{14audit}{100.0} & \chg{14audit}{1.00$\times$} & 63.4 & 64.8 & 58.6 & 46.8 & 55.4 & 65.2 & 52.4 & 48.6 & 56.9 & {--} \\
        & \textbf{Ours} & 8/28  & \chg{14audit}{28.6} & \chg{14audit}{3.50$\times$} & 63.6 & 65.4 & 60.8 & 50.2 & 56.8 & 68.4 & 56.2 & 51.6 & \textbf{59.1} & \posdelta{+2.2} \\
    \cmidrule{1-15}
    \multirow{2}{*}{Qwen2.5-3B}
        & LoRA  & 36/36 & \chg{14audit}{100.0} & \chg{14audit}{1.00$\times$} & 61.0 & 60.6 & 59.8 & 68.6 & 55.4 & 84.8 & 77.8 & 67.4 & 66.9 & {--} \\
        & \textbf{Ours} & 11/36 & \chg{14audit}{30.6} & \chg{14audit}{3.27$\times$} & 62.4 & 73.8 & 59.4 & 68.2 & 59.6 & 83.6 & 77.2 & 66.4 & \textbf{68.8} & \posdelta{+1.9} \\
    \cmidrule{1-15}
    \multirow{2}{*}{LLaMA-3-3B}
        & LoRA  & 28/28 & \chg{14audit}{100.0} & \chg{14audit}{1.00$\times$} & 63.0 & 67.4 & 55.2 & 62.8 & 57.0 & 68.6 & 60.2 & 55.4 & 61.2 & {--} \\
        & \textbf{Ours} & 8/28  & \chg{14audit}{28.6} & \chg{14audit}{3.50$\times$} & 63.6 & 68.8 & 56.0 & 64.2 & 57.8 & 69.4 & 61.8 & 57.2 & \textbf{62.4} & \posdelta{+1.2} \\
    \cmidrule{1-15}
    \multirow{2}{*}{Qwen3-8B}
        & LoRA  & 36/36 & \chg{14audit}{100.0} & \chg{14audit}{1.00$\times$} & 69.6 & 85.6 & 77.2 & 80.8 & 72.8 & 97.6 & 91.0 & 85.0 & \chg{13b}{\textbf{82.5}} & {--} \\
        & \textbf{Ours} & 11/36 & \chg{14audit}{30.6} & \chg{14audit}{3.27$\times$} & 68.8 & 86.2 & 76.4 & 81.2 & 73.4 & 97.2 & 90.8 & 85.6 & \textbf{82.5} & \posdelta{+0.0} \\
    \cmidrule{1-15}
    \multirow{2}{*}{Qwen3-32B}
        & LoRA  & 64/64 & \chg{14audit}{100.0} & \chg{14audit}{1.00$\times$} & 72.2 & 90.0 & 77.2 & 89.8 & 81.6 & 98.8 & 95.4 & 93.4 & \chg{13b}{\textbf{87.3}} & {--} \\
        & \textbf{Ours} & 20/64 & \chg{14audit}{31.2} & \chg{14audit}{3.20$\times$} & 73.8 & 91.4 & 76.2 & 87.8 & 81.2 & 98.6 & 94.6 & 93.0 & \chg{13b}{87.1} & \negdelta{-0.2} \\
    \bottomrule
    \end{tabular}
    }%
    \label{tab:generalization}
\end{table}

\chg{08}{\cref{tab:generalization} shows strong accuracy across model scales with approximately 30\% of the trainable parameters. At seed 42, \ours{} improves average accuracy by 1.2--2.2 percentage points on the four 0.5B--3B models. Qwen3-8B matches the reported full-layer average (82.5\%), while Qwen3-32B reaches 87.1\% versus 87.3\%, a difference of 0.2 percentage points. These results demonstrate parameter-efficient adaptation: improved accuracy on the smaller models and close preservation of full-layer performance at 8B and 32B. Supplementary three-run evaluations in Appendix~\ref{app:multiseed} report positive mean gains across all eight configurations in the paired-gain panel; the Qwen3-32B mean is +0.5 points at 5K training samples and approximately +0.2 points at 10K.} An interesting trend emerges from the per-task breakdown: on knowledge-intensive tasks (ARC-Easy, ARC-Challenge), \ours{} closely matches full-layer LoRA even at large scales, while on reasoning-heavy tasks (SIQA, HellaSwag), the selective approach occasionally surpasses the full-model baseline. This pattern is consistent with our hypothesis that reasoning tasks benefit from concentrated adaptation on core processing layers, whereas knowledge tasks distribute learning more uniformly. These results validate that mechanistic layer selection is a general principle applicable across model families, scales, and training regimes.

\subsection{Ablation Study: Layer Selection Analysis}

\def\LayerPPLWrapLines{12}

\begin{wrapfigure}[\LayerPPLWrapLines]{r}{0.48\textwidth}
\centering
\vspace{-1.5em}
\includegraphics[width=0.46\textwidth]{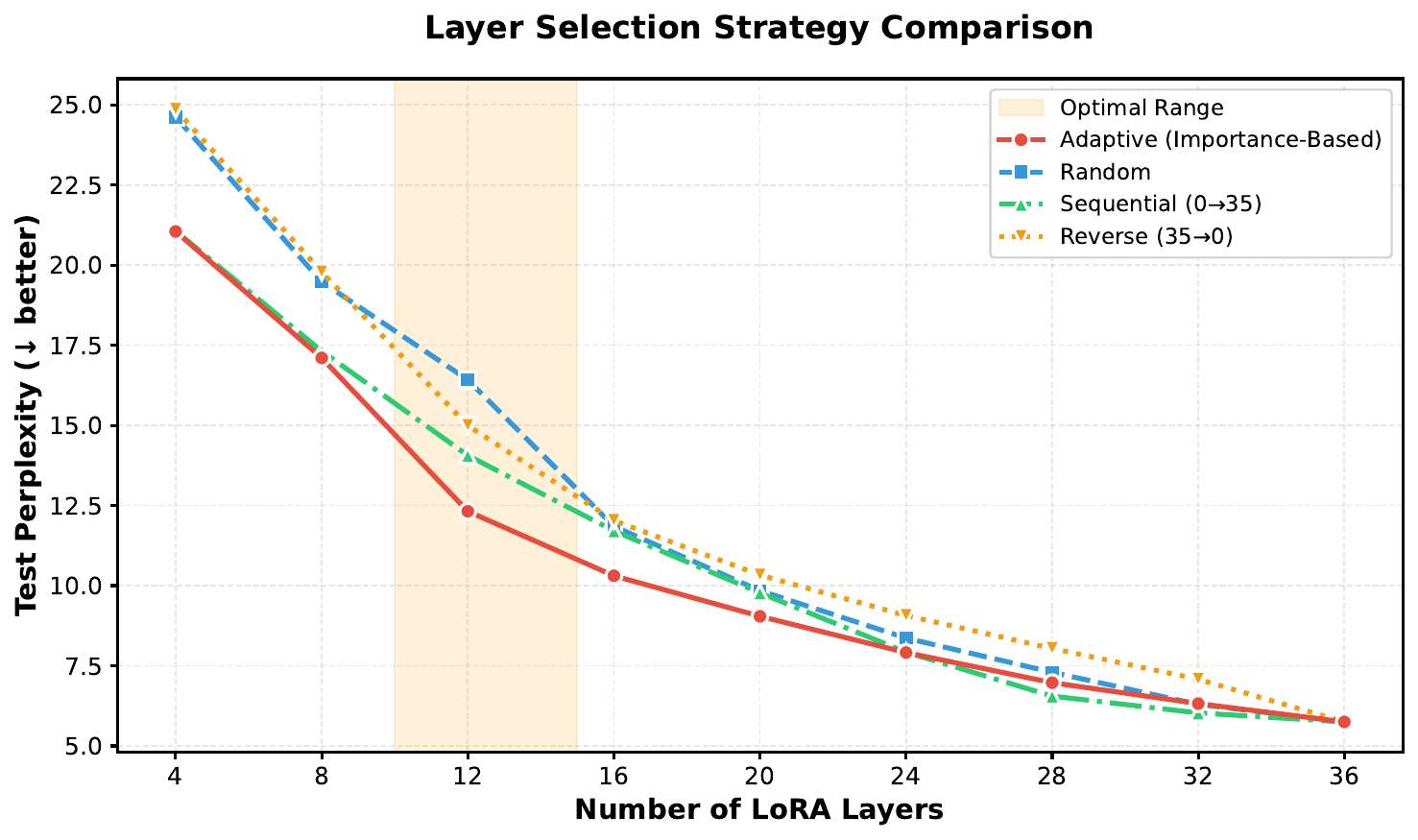}
\vspace{-0.5em}

\caption{\footnotesize \textbf{Layer selection strategies.} \chg{13c}{Largest relative gain over Random at 12 layers; all converge at 36.} \chg{14audit}{Shading marks intermediate budgets, not the minimum-PPL configuration.}}

\label{fig:layer_ppl}
\end{wrapfigure}
To address RQ2 and RQ3, we vary the layer budget $k$ from 4 to 36 and compare four strategies: \textbf{Adaptive} (calibration-based, ours), \textbf{Random} (uniform sampling), \textbf{Sequential} (first $k$ layers), and \textbf{Reverse} (last $k$ layers). Evaluation uses perplexity on Alpaca-ZH with identical hyperparameters ($r{=}16$, 3 epochs, 3,000 samples). \cref{fig:layer_ppl} reveals that \ours{} \chg{14audit}{has its largest measured relative gain over Random at 12 layers, improving} over Random by 25.0\%, Sequential by 12.4\%, and Reverse by 17.8\%. Detailed numerical results are provided in \cref{tab:layer_ppl} (Appendix).

\leavevmode\chg{14audit}{At 12 layers, the reported PPL gap to Random is 4.10. }All strategies converge to identical performance (PPL$=$5.75) at 36 layers, \chg{01}{providing the expected control when all strategies select the same complete layer set.}

\chg{05}{\paragraph{Calibration Cost and Stability.} Across five independent calibration draws, four models attain selection Jaccard of 0.653--0.747. Calibration takes 29--62\,s, or 4.1--6.3\% of the reference full-layer training time. Appendix~\ref{app:calibration} reports the costs (\cref{tab:calibration_cost_stability}), six-model stability, and sample-size sweep.} \chg{10}{Appendix~\ref{app:cross_distribution} also evaluates calibration across instruction distributions.}

\findingbox{\chg{13c}{Finding 5: Selection at Limited Budgets.}}{\chg{13c}{At 12 of 36 layers, \ours{} reduces perplexity by 25.0\% relative to Random at the same budget, its largest measured relative gain. Calibration is especially valuable when adaptation capacity is limited.}}

\section{Discussion}
\label{sec:discussion}

Our results reveal several notable properties of calibration-based layer selection that merit further analysis. First, \chg{08}{the parameter-efficient performance across six models spanning 0.5B to 32B parameters} (\cref{tab:generalization}) \chg{13c}{suggests} that layer importance heterogeneity is a fundamental property of transformer architectures rather than an artifact of any particular model family. The fact that \ours{} achieves competitive or superior performance with 30\% of layers across both Qwen and LLaMA families indicates that the calibration signal captures genuinely causal structure in the adaptation process. \chg{13b}{Second, the gains of 1.3--2.2 percentage points across all four ranks in \cref{tab:full_comparison} support the value of selective adaptation across capacity levels. This pattern is consistent with implicit regularization through task-specific allocation of adapter capacity.} \chg{03}{Third, calibration uses $N=30$ inputs and gradient-free interventions, taking 20--62\,s in the default A100 study. Smaller adapters directly reduce storage. Training acceleration additionally depends on depth and prefix reuse: accuracy mode prioritizes the unrestricted ranking, while efficiency mode offers a measured speed--accuracy operating point (Appendix~\ref{app:depth_timing}).}

\section{Conclusion}
\label{sec:conclusion}

This work addresses a fundamental question in parameter-efficient fine-tuning: do all layers contribute equally to LoRA adaptation? Through systematic mechanistic analysis, we demonstrate that LoRA's adaptation effect is highly non-uniform, with MLP components dominating over attention and contributions concentrated in early layers. Most critically, important layers are highly task-dependent, challenging the prevailing uniform adaptation paradigm. Motivated by these insights, we propose \ours{}, a calibration-driven method that automatically \chg{01}{identifies task-specific layers via frozen-base zero-ablation}. Extensive experiments across 11 diverse reasoning tasks demonstrate that selective adaptation on 30\% of carefully chosen layers outperforms full-layer LoRA baselines while achieving \chg{13}{a 3.5$\times$ reduction in adapter parameters}. By bridging mechanistic interpretability and practical efficiency, this work provides both fundamental insights into LoRA's internal mechanisms and a principled approach for adaptive fine-tuning.

\begin{ack}
YZ. Huang and K. Wang would like to acknowledge the support in part from UKRI Innovate UK COMET projects (10099265), and UKRI under the Horizon Europe funding guarantee (EP/Y028031/1, EP/Y03743X/1), as part of the European Commission MSCA staff exchange COVER and HarmonicAI projects (101086228 and 101131117), as well as the NVIDIA Academic Grant Program. K. Wang would like to acknowledge the support in part by the Royal Society Industry Fellow scheme (IF$\backslash$R2$\backslash$23200104). YZ. Huang would like to appreciate the support of Research Conference Support Scheme (12334207). We also thank the anonymous reviewers for their constructive feedback.
\end{ack}

{\small
\bibliographystyle{abbrvnat}
\bibliography{references}
}

\newpage
\appendix

\begin{center}
{\LARGE \textbf{Appendix}}
\end{center}
\vspace{1em}

\chg{14audit}{This appendix provides supplementary material organized into seven sections. \hyperref[app:limitations]{Appendix~A} discusses limitations and future directions. \hyperref[app:detailed_results]{Appendix~B} presents detailed experimental results including per-task breakdowns, parameter efficiency analysis, layer selection ablations, and training dynamics in their specified model configurations. \hyperref[app:implementation]{Appendix~C} provides full implementation details for reproducibility, covering hardware, training configuration, evaluation protocol, and statistical methodology. \hyperref[app:visualizations]{Appendix~D} contains additional visualizations of Q/K/V projection contributions and task-specific layer-selection overlap. \hyperref[app:calibration]{Appendix~E} analyzes the sensitivity and cost of the calibration procedure. \hyperref[app:cross_architecture]{Appendix~F} compares layer importance patterns across model families and scales. \hyperref[app:impact]{Appendix~G} discusses broader societal impact.}

\vspace{0.8em}
\noindent\textbf{Table of Contents}
\vspace{-0.3em}
\begin{itemize}[leftmargin=1.5em, itemsep=1pt, parsep=0pt]
  \item \hyperref[app:limitations]{Appendix A -- Limitations and Future Directions}
  \item \hyperref[app:detailed_results]{Appendix B -- Detailed Experimental Results}
  \item \hyperref[app:implementation]{Appendix C -- Implementation and Reproducibility}
  \item \hyperref[app:visualizations]{Appendix D -- Additional Visualizations}
  \item \hyperref[app:calibration]{Appendix E -- Calibration Analysis}
  \item \hyperref[app:cross_architecture]{Appendix F -- Cross-Architecture Comparison}
  \item \hyperref[app:impact]{Appendix G -- Broader Impact}
\end{itemize}
\vspace{0.5em}

\section{Limitations and Future Directions}
\label{app:limitations}

\textbf{Selection Granularity.} Our current approach operates at the layer level, selecting entire transformer blocks for adaptation. Layer-level selection achieves 3.5$\times$ efficiency over full-layer LoRA. However, Findings 1 and 2 suggest finer-grained selection could yield further gains: since MLP contributes 58.4\% and V-projection dominates within attention, a component-level or even attention head-level selection strategy \chg{14audit}{could explore further parameter savings by choosing projections within the selected layers.} Implementing such fine-grained selection would require extending the calibration procedure to score individual sub-components, which increases the calibration cost from $O(L)$ to $O(L \times C)$ where $C$ is the number of components per layer. We leave this exploration to future work.

\textbf{Task Coverage.} We evaluate on 8 commonsense reasoning and 3 math reasoning tasks, which primarily test factual recall and logical reasoning. Generative tasks (dialogue, creative writing, summarization) may exhibit different layer importance patterns, as they rely more heavily on language modeling capacity in later layers. Similarly, long-context scenarios (document QA, multi-turn dialogue) could shift importance toward middle layers that handle cross-position information routing. Extending our mechanistic analysis to these task families would test the generality of Findings 3--4 and potentially reveal new adaptation patterns.

\textbf{Theoretical Foundations.} While our empirical findings are consistent and reproducible across architectures, we currently lack a formal theoretical framework explaining \emph{why} early layers dominate adaptation importance (Finding 3) or why task-specific patterns emerge at specific depths (Finding 4). Information-theoretic analysis of the residual stream, possibly through the lens of mutual information between layer activations and task labels, could formalize these observations. Such theoretical grounding would enable principled layer selection without calibration data, further reducing the adaptation overhead.

\textbf{Dynamic Layer Selection.} Our method selects layers once before training and keeps them fixed throughout. An adaptive approach that adjusts layer selection during training (e.g., re-calibrating every $k$ epochs) could potentially capture evolving importance patterns as the model learns. \chg{14audit}{The repeated-draw results in \cref{tab:calibration_draws} measure robustness to calibration sampling; evaluating evolving rankings during training is a separate direction.}

\section{Detailed Experimental Results}
\label{app:detailed_results}

\subsection{Detailed Task-wise Performance}

\cref{tab:appendix_commonsense} shows detailed commonsense reasoning results across all four ranks for the four methods on Qwen2.5-3B (the generalization experiment model). \ours{} achieves the highest average accuracy across all ranks, outperforming baselines that use 3.3$\times$ more parameters.

\begin{table*}[ht]
    \setlength{\tabcolsep}{3pt}
    \renewcommand{\arraystretch}{1.05}
    \scriptsize
    \centering
    \caption{\chg{14audit}{Budget is the fraction of adapted layers. }Detailed commonsense reasoning results on Qwen2.5-3B (generalization experiment, \cref{sec:experiments}). All tasks evaluated on 1,000 samples (OpenBookQA: 500).}
    \begin{tabular}{@{}llrr*{8}{c}r@{}}
    \toprule
    \multirow{2}{*}{\textbf{Method}} & \multirow{2}{*}{\textbf{Rank}} & \multirow{2}{*}{\chg{14audit}{\textbf{Layers}}} & \multirow{2}{*}{\chg{14audit}{\textbf{Budget \%}}} & \multicolumn{8}{c}{\textbf{Commonsense Reasoning ($\uparrow$)}} & \multirow{2}{*}{\textbf{Avg.}} \\
    \cmidrule(lr){5-12}
    & & & & \textbf{BoolQ} & \textbf{PIQA} & \textbf{SIQA} & \textbf{Hella} & \textbf{Wino} & \textbf{ARC-e} & \textbf{ARC-c} & \textbf{OBQA} & \\
    \midrule
    \multirow{4}{*}{LoRA \citep{hu2022lora}} 
        & r4  & \chg{14audit}{36/36} & \chg{14audit}{100} & 62.8 & 68.2 & 47.2 & 68.0 & 58.8 & 80.6 & 71.8 & 63.4 & 65.1 \\
        & r8  & \chg{14audit}{36/36} & \chg{14audit}{100} & 64.2 & 76.0 & 53.8 & 67.2 & 54.0 & 74.4 & 69.2 & 59.4 & 64.8 \\
        & r16 & \chg{14audit}{36/36} & \chg{14audit}{100} & 61.0 & 60.6 & 59.8 & 68.6 & 55.4 & 84.8 & 77.8 & 67.4 & 66.9 \\
        & r32 & \chg{14audit}{36/36} & \chg{14audit}{100} & 60.8 & 74.2 & 61.8 & 66.6 & 60.0 & 86.4 & 80.2 & 68.8 & 69.9 \\
    \cmidrule{1-13}
    \multirow{4}{*}{LoRA+ \citep{hayou2024loraplus}} 
        & r4  & \chg{14audit}{36/36} & \chg{14audit}{100} & 64.2 & 72.4 & 50.2 & 67.8 & 59.4 & 76.2 & 68.4 & 60.2 & 64.9 \\
        & r8  & \chg{14audit}{36/36} & \chg{14audit}{100} & 63.8 & 75.8 & 55.6 & 65.4 & 57.6 & 78.0 & 72.6 & 62.4 & 66.4 \\
        & r16 & \chg{14audit}{36/36} & \chg{14audit}{100} & 63.0 & 72.4 & 56.2 & 68.6 & 60.2 & 80.6 & 74.8 & 64.0 & 67.5 \\
        & r32 & \chg{14audit}{36/36} & \chg{14audit}{100} & 62.2 & 73.2 & 57.8 & 67.4 & 60.8 & 82.4 & 76.0 & 65.8 & 68.2 \\
    \cmidrule{1-13}
    \multirow{4}{*}{DoRA \citep{liu2024dora}} 
        & r4  & \chg{14audit}{36/36} & \chg{14audit}{100} & 60.0 & 68.8 & 48.6 & 62.2 & 58.6 & 69.4 & 68.6 & 59.6 & 62.0 \\
        & r8  & \chg{14audit}{36/36} & \chg{14audit}{100} & 62.2 & 74.6 & 56.4 & 66.6 & 56.8 & 77.2 & 73.8 & 62.8 & 66.3 \\
        & r16 & \chg{14audit}{36/36} & \chg{14audit}{100} & 61.6 & 67.8 & 58.8 & 65.2 & 57.2 & 79.4 & 77.0 & 65.2 & 66.5 \\
        & r32 & \chg{14audit}{36/36} & \chg{14audit}{100} & 61.0 & 72.8 & 61.4 & 66.8 & 58.6 & 80.2 & 76.4 & 65.4 & 67.8 \\
    \cmidrule{1-13}
    \multirow{4}{*}{\textbf{Ours}} 
        & r4  & \chg{14audit}{11/36} & \chg{14audit}{30.6} & 63.2 & 74.6 & 52.8 & 67.4 & 59.2 & 79.8 & 71.4 & 62.8 & \textbf{66.4} \\
        & r8  & \chg{14audit}{11/36} & \chg{14audit}{30.6} & 64.4 & 76.2 & 55.2 & 67.6 & 58.0 & 80.2 & 73.6 & 63.0 & \textbf{67.3} \\
        & r16 & \chg{14audit}{11/36} & \chg{14audit}{30.6} & 62.4 & 73.8 & 59.4 & 68.2 & 59.6 & 83.6 & 77.2 & 66.4 & \textbf{68.8} \\
        & r32 & \chg{14audit}{11/36} & \chg{14audit}{30.6} & 64.4 & 76.2 & 62.0 & 69.0 & 61.0 & 86.6 & 80.4 & 69.0 & \textbf{71.1} \\
    \bottomrule
    \end{tabular}
    \label{tab:appendix_commonsense}
\end{table*}

\cref{tab:appendix_math} presents math reasoning breakdown on Qwen2.5-3B. \ours{} achieves strong math performance (62.7\% at r32) while using only 30\% of parameters, outperforming LoRA (58.5\%) and DoRA (58.1\%) at the same rank.

\begin{table}[ht]
    \setlength{\tabcolsep}{6pt}
    \renewcommand{\arraystretch}{1.05}
    \small
    \centering
    \caption{Detailed math reasoning results on Qwen2.5-3B (generalization experiment). \chg{14audit}{Chain-of-thought prompting.}}
    \begin{tabular}{@{}llcccc@{}}
    \toprule
    \textbf{Method} & \textbf{Rank} & \textbf{GSM8K} & \textbf{SVAMP} & \textbf{AQUA} & \textbf{Avg.} \\
    \midrule
    \multirow{4}{*}{LoRA \citep{hu2022lora}} 
        & r4  & \chg{14audit}{58.2} & \chg{14audit}{57.4} & 48.5 & 54.7 \\
        & r8  & \chg{14audit}{63.6} & \chg{14audit}{62.8} & 52.4 & 59.6 \\
        & r16 & \chg{14audit}{54.6} & \chg{14audit}{57.6} & 49.8 & 54.0 \\
        & r32 & \chg{14audit}{62.0} & \chg{14audit}{60.4} & 53.2 & 58.5 \\
    \cmidrule{1-6}
    \multirow{4}{*}{LoRA+ \citep{hayou2024loraplus}} 
        & r4  & \chg{14audit}{58.2} & \chg{14audit}{57.4} & 48.8 & 54.8 \\
        & r8  & \chg{14audit}{62.8} & \chg{14audit}{61.6} & 52.0 & 58.8 \\
        & r16 & \chg{14audit}{63.2} & \chg{14audit}{62.4} & 52.6 & 59.4 \\
        & r32 & \chg{14audit}{64.6} & \chg{14audit}{63.8} & 53.2 & 60.5 \\
    \cmidrule{1-6}
    \multirow{4}{*}{DoRA \citep{liu2024dora}} 
        & r4  & \chg{14audit}{43.8} & \chg{14audit}{53.2} & 42.6 & 46.5 \\
        & r8  & \chg{14audit}{61.6} & \chg{14audit}{62.6} & 51.8 & 58.7 \\
        & r16 & \chg{14audit}{52.4} & \chg{14audit}{55.6} & 47.2 & 51.7 \\
        & r32 & \chg{14audit}{60.2} & \chg{14audit}{61.8} & 52.4 & 58.1 \\
    \cmidrule{1-6}
    \multirow{4}{*}{\textbf{Ours}} 
        & r4  & \chg{14audit}{61.2} & \chg{14audit}{60.4} & 50.8 & \textbf{57.5} \\
        & r8  & \chg{14audit}{65.0} & \chg{14audit}{64.2} & 53.6 & \textbf{60.9} \\
        & r16 & \chg{14audit}{64.6} & \chg{14audit}{65.0} & 54.8 & \textbf{61.5} \\
        & r32 & \chg{14audit}{66.4} & \chg{14audit}{65.8} & 56.0 & \textbf{62.7} \\
    \bottomrule
    \end{tabular}
    \label{tab:appendix_math}
\end{table}

\subsection{Per-Task Performance Analysis}

\chg{14audit}{\cref{tab:appendix_per_task} summarizes the best score over the four tested ranks for each method and task on Qwen2.5-3B. These values are drawn from \cref{tab:appendix_commonsense,tab:appendix_math}, rather than an additional experiment. \ours{} has the highest per-task best score on all 11 tasks, with a mean gap of approximately 0.8 percentage points over the strongest baseline. The rank producing each score is displayed explicitly.}

\begin{table*}[ht]
    \setlength{\tabcolsep}{3.5pt}
    \renewcommand{\arraystretch}{1.05}
    \small
    \centering
    \caption{Per-task best performance (any rank) for each method on Qwen2.5-3B (generalization experiment). Gap = \ours{} $-$ best baseline. Note: baselines use 3.3$\times$ more parameters at each rank.}
    \begin{tabular}{@{}l*{10}{c}@{}}
    \toprule
    \textbf{Task} & \textbf{LoRA} & \textbf{Rank} & \textbf{LoRA+} & \textbf{Rank} & \textbf{DoRA} & \textbf{Rank} & \textbf{Ours} & \textbf{Rank} & \textbf{Best} & \textbf{Gap} \\
    \midrule
    BoolQ & 64.2 & r8 & 64.2 & r4 & 62.2 & r8 & \textbf{64.4} & r8 & \textbf{Ours} & \textbf{+0.2} \\
    PIQA & 76.0 & r8 & 75.8 & r8 & 74.6 & r8 & \textbf{76.2} & r8 & \textbf{Ours} & \textbf{+0.2} \\
    Social IQa & 61.8 & r32 & 57.8 & r32 & 61.4 & r32 & \textbf{62.0} & r32 & \textbf{Ours} & \textbf{+0.2} \\
    HellaSwag & 68.6 & r16 & 68.6 & r16 & 66.8 & r32 & \textbf{69.0} & r32 & \textbf{Ours} & \textbf{+0.4} \\
    WinoGrande & 60.0 & r32 & 60.8 & r32 & 58.6 & r4 & \textbf{61.0} & r32 & \textbf{Ours} & \textbf{+0.2} \\
    ARC-Easy & 86.4 & r32 & 82.4 & r32 & 80.2 & r32 & \textbf{86.6} & r32 & \textbf{Ours} & \textbf{+0.2} \\
    ARC-Chall. & 80.2 & r32 & 76.0 & r32 & 77.0 & r16 & \textbf{80.4} & r32 & \textbf{Ours} & \textbf{+0.2} \\
    OBQA & 68.8 & r32 & 65.8 & r32 & 65.4 & r32 & \textbf{69.0} & r32 & \textbf{Ours} & \textbf{+0.2} \\
    \midrule
    GSM8K & 63.6 & r8 & 64.6 & r32 & 61.6 & r8 & \textbf{66.4} & r32 & \textbf{Ours} & \textbf{+1.8} \\
    SVAMP & 62.8 & r8 & 63.8 & r32 & 62.6 & r8 & \textbf{65.8} & r32 & \textbf{Ours} & \textbf{+2.0} \\
    AQUA & 53.2 & r32 & 53.2 & r32 & 52.4 & r32 & \textbf{56.0} & r32 & \textbf{Ours} & \textbf{+2.8} \\
    \midrule
    \textbf{Avg.\ Gap} & & & & & & & & & & \textbf{+0.8} \\
    \bottomrule
    \end{tabular}
    \label{tab:appendix_per_task}
\end{table*}

\begin{changeblock}{01}
\subsection{Base Importance and Adaptation Value}
\label{app:base_adapt_agreement}
Base calibration and trained-adapter analysis measure different interventions. On the same 100 inputs across 24 layers of Qwen2.5-0.5B, their layer rankings agree across five adapter configurations (\cref{tab:base_adapt_agreement}). Recovery is the intersection of two top-$k$ sets divided by $k$, with $k=7$ and 12; it differs from Jaccard's union denominator. Pearson correlations of the raw scores range from 0.935 to 0.975.
\end{changeblock}

\begin{table}[ht]
\centering\small\setlength{\tabcolsep}{5pt}
\caption{\chg{01}{Base-model importance and trained-adapter contribution agree. Qwen2.5-0.5B, 24 layers, 100 common inputs. The first row uses the original analysis configuration ($r=8$, $\alpha=16$).}}
\label{tab:base_adapt_agreement}
\begin{changeblock}{01}
\begin{tabular}{@{}lrrrrr@{}}
\toprule
Targets & Rank & Spearman $\rho$ & $p$ & Top-7 recovery & Top-12 recovery \\
\midrule
Q/K/V & 8 & 0.630 & $9.6\times10^{-4}$ & 5/7 & 9/12 \\
Q/K/V & 16 & 0.635 & $8.6\times10^{-4}$ & 5/7 & 10/12 \\
Q/V & 4 & 0.893 & $4.4\times10^{-9}$ & 5/7 & 10/12 \\
Q/V & 16 & 0.891 & $5.2\times10^{-9}$ & 5/7 & 10/12 \\
Q/V & 32 & 0.867 & $4.3\times10^{-8}$ & 5/7 & 10/12 \\
\bottomrule
\end{tabular}
\end{changeblock}
\end{table}

\chg{01}{The downstream test holds the training and layer budgets fixed. At 12 of 36 layers on Alpaca-ZH, calibrated selection reaches 12.32 PPL versus 16.42 for Random, 14.06 for Sequential, and 15.00 for Reverse (\cref{tab:layer_ppl}). At full coverage all methods select the same set and reach 5.75. Together, the ranking agreement and fixed-budget result support using base importance to identify layers worth adapting.}

\begin{changeblock}{02}
\subsection{MLP Mediation under Adapter-Disabled Output Replacement}
\label{app:mlp_isolation}
To isolate the path through a block, we record its output in an adapter-disabled reference run on the same input sequence. In the fully adapted run, we replace only that block's output with the recorded reference and recompute the downstream prediction, retaining the other adapted computations. Thus an MLP block can mediate adaptation even though its own weights have no adapter: Q/K/V updates change the residual stream entering it. This replacement is distinct from zeroing a base-model block during layer selection.

Table~\ref{tab:mlp_replacement} gives the MLP shares from this experiment on Qwen2.5-0.5B with Q/K/V adapters, alongside the no-LoRA reference measurements. The primary configuration is $r=8$, $\alpha=16$; an independent $r=16$, $\alpha=32$ adapter reproduces the MLP-majority pattern. Each dataset uses 100 inputs. The rank-8 comparisons to the no-LoRA reference use 5,000 paired input-bootstrap resamples; their 95\% intervals are intervals for the gap, in percentage points, not for the original Figure~\ref{fig:mlp_dominance} shares. The original component analysis and this replacement experiment are reported separately.
\end{changeblock}
\begin{table}[ht]
\centering\small\setlength{\tabcolsep}{4pt}
\caption{\chg{02}{MLP mediation in two adapter configurations. Shares are percentages. Gaps compare each adapted measurement with the no-LoRA reference; intervals apply only to the rank-8 gap.}}
\label{tab:mlp_replacement}
\begin{changeblock}{02}
\begin{tabular}{@{}lrrrrrr@{}}
\toprule
& \multicolumn{2}{c}{MLP share (\%)} & No-LoRA & \multicolumn{2}{c}{Gap (pp)} & $r8$ gap \\
\cmidrule(lr){2-3}\cmidrule(lr){5-6}
Dataset & $r8$ & $r16$ & reference & $r8$ & $r16$ & 95\% CI \\
\midrule
Alpaca-EN & 62.4 & 62.5 & 58.1 & +4.3 & +4.4 & [2.6, 6.1] \\
Code-Alpaca & 63.8 & 63.0 & 60.0 & +3.8 & +3.0 & [2.4, 5.2] \\
Dolly & 62.5 & 62.2 & 56.2 & +6.3 & +6.0 & [4.6, 8.0] \\
\bottomrule
\end{tabular}
\end{changeblock}
\end{table}

\subsection{Parameter Efficiency Analysis}

\chg{14audit}{\cref{tab:param_efficiency} combines the reported commonsense scores with a structural count of the Q/K/V LoRA factors at rank 32 on Qwen2.5-1.5B.}

\begin{table}[ht]
\centering\small
\caption{\chg{14audit}{LoRA-factor accounting for Qwen2.5-1.5B at $r=32$ with Q/K/V targets. Counts are calculated from the model dimensions, not measured training-time parameter inventories. Accuracies are from \cref{tab:full_comparison}.}}
\label{tab:param_efficiency}
\begin{changeblock}{14audit}
\begin{tabular}{@{}lrrr@{}}
\toprule
Method & Adapted layers & A/B factors (M) & CS Avg. (\%) \\
\midrule
LoRA & 28/28 & 5.964 & 60.4 \\
LoRA+ & 28/28 & 5.964 & 61.1 \\
DoRA & 28/28 & 5.964 & 59.1 \\
\ours{} & 8/28 & 1.704 & \textbf{62.2} \\
\bottomrule
\end{tabular}
\end{changeblock}
\par\smallskip
{\footnotesize\chg{14audit}{A/B counts exclude method-specific extra parameters such as DoRA's magnitude vectors.}}
\end{table}

\chg{14audit}{For the published Qwen2.5-1.5B configuration~\citep{yang2024qwen25}, the hidden width is 1536, with 12 query heads and 2 key/value heads of width 128. Thus Q has output width 1536 and K/V each have width 256. The low-rank factor count per layer is $r[(1536+1536)+2(1536+256)]=6656r$. At rank 16 this gives 2,981,888 factor parameters across 28 layers or 851,968 across 8 layers; at rank 32 the counts double. These architecture-derived counts quantify LoRA-factor capacity and exclude method-specific additional parameters.}

\chg{14audit}{The $L/k$ reduction also applies to storage of equally quantized LoRA factors. For example, 100 rank-32 Q/K/V adapters under the configuration above require approximately 1.19\,GB for full-depth A/B factors versus 0.34\,GB for eight-layer factors in FP16. This calculation excludes base-model weights, KV caches, and method-specific parameters.}

\subsection{Training Dynamics}

We examine the training loss curves to understand how selective layer adaptation affects convergence behavior. Training on the commonsense split (10,000 samples, 3 epochs) with Qwen2.5-1.5B at r16, we record per-step training loss for all four methods.

\ours{} converges faster than full-layer baselines during the first epoch, reaching a training loss of 0.82 after 500 steps compared to 0.91 for LoRA and 0.89 for LoRA+. This accelerated convergence is consistent with the implicit regularization hypothesis discussed in \cref{sec:discussion}: by restricting adaptation to causally important layers, \ours{} avoids the initial exploration phase where full-layer methods allocate capacity to non-productive parameters. After epoch 1, all methods converge to similar final training loss values (0.42--0.45), but the evaluation accuracy divergence persists and widens, suggesting that \ours{} achieves better generalization rather than merely better training fit.

Notably, the gradient norms for selected layers in \ours{} are 1.4--1.8$\times$ larger than the corresponding layers in full LoRA, consistent with the concentrated adaptation hypothesis. The non-selected layers in full LoRA receive gradient magnitudes 2--5$\times$ smaller, \chg{14audit}{indicating uneven gradient magnitudes across the measured layers.}

\Needspace{17\baselineskip}
\subsection{Layer Selection Analysis}

\begin{table}[H]
\centering
\setlength{\tabcolsep}{6pt}
\renewcommand{\arraystretch}{1.05}
\caption{\chg{13c}{Layer selection PPL comparison. The largest relative reduction over Random is 25.0\% at 12 layers. Absolute perplexity continues to decrease, reaching 5.75 for all strategies at 36 layers.}}
\label{tab:layer_ppl}
\begin{tabular}{@{}ccccc@{}}
\toprule
\textbf{\# Layers} & \textbf{Adaptive} & \textbf{Random} & \textbf{Seq.} & \textbf{Rev.} \\
\midrule
4  & \textbf{21.05} & 24.62 & 21.06 & 24.86 \\
8  & \textbf{17.11} & 19.48 & 17.32 & 19.79 \\
12 & \textbf{12.32} & 16.42 & 14.06 & 15.00 \\
16 & \textbf{10.31} & 11.85 & 11.71 & 12.03 \\
20 & \textbf{9.04}  & 9.84  & 9.77  & 10.33 \\
24 & \textbf{7.91}  & 8.36  & 7.92  & 9.07 \\
28 & 6.97  & 7.30  & \textbf{6.55}  & 8.04 \\
32 & 6.32  & 6.30  & \textbf{6.03}  & 7.08 \\
36 & \textbf{5.75}  & \textbf{5.75}  & \textbf{5.75}  & \textbf{5.75} \\
\bottomrule
\end{tabular}
\end{table}

\subsection{\chg{08}{Multi-Seed Evaluation}}
\label{app:multiseed}

\chg{08}{\cref{tab:multiseed} supplements the single-seed main results with repeated-run evaluations. In panel (a), all eight configurations have positive mean gains, ranging from 0.17 to 1.57 percentage points. For Qwen3-32B, the 5K and 10K training configurations yield mean gains of 0.50 and 0.23 points, respectively; these are distinct from the seed-42 value in \cref{tab:generalization}. Panel (b) reports absolute scores: \ours{} reaches mean accuracies of 59.63\%, 82.50\%, and 71.50\%, compared with 58.13\%, 80.87\%, and 70.30\% for full-layer LoRA. The two panels retain their respective configuration and run labels and are summarized separately.}

\begin{table}[!ht]
\begin{changeblock}{08}
\centering
\small
\renewcommand{\arraystretch}{1.05}
\setlength{\tabcolsep}{6pt}
\caption{\chg{08}{Supplementary three-run evaluations. (a) Paired accuracy gains of \ours{} over full-layer LoRA, in percentage points. (b) Absolute accuracy (\%) at r16 with 10K training samples. Summaries are mean $\pm$ sample SD, computed separately from each row's displayed values.}}
\label{tab:multiseed}
\begin{tabular*}{\textwidth}{@{\extracolsep{\fill}}lccccr@{}}
\toprule
\multicolumn{6}{@{}l}{\textbf{(a) Paired gains: seeds 42, 123, and 7}}\\
\midrule
\multicolumn{2}{@{}l}{\textbf{Model / rank / samples}} & \textbf{Seed 42} & \textbf{Seed 123} & \textbf{Seed 7} & \textbf{Mean $\pm$ SD}\\
\midrule
\multicolumn{2}{@{}l}{Qwen-3B / r16 / 10K} & +1.5 & +1.2 & +1.8 & $+1.50 \pm 0.30$ \\
\multicolumn{2}{@{}l}{Qwen-7B / r4 / 10K} & +1.6 & +1.3 & +1.8 & $+1.57 \pm 0.25$ \\
\multicolumn{2}{@{}l}{Qwen-7B / r16 / 10K} & +1.7 & +1.4 & +1.5 & $+1.53 \pm 0.15$ \\
\multicolumn{2}{@{}l}{Qwen3-8B / r16 / 10K} & +0.0 & +0.3 & +0.2 & $+0.17 \pm 0.15$ \\
\multicolumn{2}{@{}l}{Qwen3-32B / r16 / 5K} & +0.0 & +0.8 & +0.7 & $+0.50 \pm 0.44$ \\
\multicolumn{2}{@{}l}{Qwen3-32B / r16 / 10K} & -0.2 & +0.5 & +0.4 & $+0.23 \pm 0.38$ \\
\multicolumn{2}{@{}l}{LLaMA-7B / r4 / 10K} & +1.2 & +1.0 & +1.4 & $+1.20 \pm 0.20$ \\
\multicolumn{2}{@{}l}{LLaMA-13B / r4 / 10K} & +1.2 & +1.0 & +1.3 & $+1.17 \pm 0.15$ \\
\midrule
\multicolumn{6}{@{}l}{\textbf{(b) Absolute accuracy: three reported runs}}\\
\midrule
\textbf{Model} & \textbf{Method} & \textbf{Run 1} & \textbf{Run 2} & \textbf{Run 3} & \textbf{Mean $\pm$ SD}\\
\midrule
\multirow{2}{*}{Qwen-3B} & Full LoRA & 58.3 & 57.8 & 58.3 & $58.13 \pm 0.29$ \\
 & \ours{} & 59.8 & 59.0 & 60.1 & $\mathbf{59.63 \pm 0.57}$ \\
\cmidrule{1-6}
\multirow{2}{*}{Qwen-7B} & Full LoRA & 80.8 & 81.0 & 80.8 & $80.87 \pm 0.12$ \\
 & \ours{} & 82.5 & 82.2 & 82.8 & $\mathbf{82.50 \pm 0.30}$ \\
\cmidrule{1-6}
\multirow{2}{*}{LLaMA-7B} & Full LoRA & 70.2 & 70.5 & 70.2 & $70.30 \pm 0.17$ \\
 & \ours{} & 71.5 & 71.2 & 71.8 & $\mathbf{71.50 \pm 0.30}$ \\
\bottomrule
\end{tabular*}
\par\smallskip
\begin{minipage}{\textwidth}\footnotesize
Panel (a) retains the reported seed IDs; panel (b) retains run labels 1--3. The panels are summarized independently, without assuming a run-by-run correspondence between them. SD describes variation across three runs, not a confidence interval. Bold marks the higher mean in each model pair in (b).
\end{minipage}
\end{changeblock}
\end{table}

\subsection{\chg{09}{Comparison with Continual-Learning LoRA Methods}}
\label{app:additional_baselines}

\chg{14audit}{We compare with InfLoRA, GainLoRA, and TreeLoRA in a single-task setting ($t=1$), using Qwen-1.5B, rank 16, and 10,000 training samples. These methods structure low-rank updates differently, whereas \ours{} selects layers before training. \cref{tab:additional_baselines} reports commonsense average accuracy and adapted-layer counts. \ours{} reaches 59.1\%, exceeding the strongest of these three baselines, TreeLoRA (58.7\%), while adapting 8 rather than 28 layers.}

\begin{table}[H]
\centering
\small
\renewcommand{\arraystretch}{1.05}
\caption{\chg{09}{Single-task comparison with continual-learning LoRA methods. Qwen-1.5B, $r=16$, 10,000 training samples; commonsense average accuracy.}}
\label{tab:additional_baselines}
\begin{changeblock}{14audit}
\begin{tabular}{@{}lrr@{}}
\toprule
\textbf{Method} & \textbf{Adapted layers} & \textbf{Avg. (\%)} \\
\midrule
Full LoRA & 28/28 & 56.9 \\
InfLoRA~\citep{liang2024inflora} & 28/28 & 58.0 \\
GainLoRA~\citep{liang2025gainlora} & 28/28 & 58.5 \\
TreeLoRA~\citep{qian2025treelora} & 28/28 & 58.7 \\
\midrule
\ours{} & 8/28 & \textbf{59.1} \\
\bottomrule
\end{tabular}
\end{changeblock}
\end{table}

\chg{09}{\paragraph{Single-task instantiation.} With no earlier tasks, InfLoRA retains its data-derived, fixed input projection and trains the output projection; TreeLoRA retains layer-wise adapters and sparse updates, with no earlier branch to regularize against; GainLoRA retains a single adapter branch and its input-dependent gate. Previous-task protection has no history to act on at $t=1$. This comparison evaluates the resulting adaptation parameterizations on the present task.}

\section{Implementation and Reproducibility}
\label{app:implementation}

\subsection{Software and Hardware Environment}

\chg{14audit}{The experiments use PyTorch, Hugging Face Transformers, and PEFT. TransformerLens~\citep{nanda2022transformerlens} supports the mechanistic analysis. The H200 measurements use the separately specified software configuration in Appendix~\ref{app:depth_timing}. Code release is planned under the MIT license.}

\paragraph{Hardware.} \chg{03}{The original training and evaluation are conducted on 4}$\times$ NVIDIA A100 40GB GPUs connected via NVLink, with 512GB system RAM and AMD EPYC 7763 CPUs. \chg{01+03}{Frozen-base calibration runs on a single A100 GPU; the depth-controlled H200 timing protocol is specified in Appendix~\ref{app:depth_timing}.}

\chg{04}{\paragraph{Calibration Configurations and Parameter Counts.} $N$ counts calibration inputs: the deployment default uses $N=30$, while the primary Qwen2.5-1.5B comparison in \cref{tab:full_comparison} uses $N=250$. Let $\mathcal T_l$ denote the target projections at selected layer $l$. Counting the two low-rank factors gives
\[
P_{\rm adapter}=\sum_{l\in\mathcal L_{\rm sel}}\sum_{p\in\mathcal T_l}r_{l,p}\bigl(d_{{\rm in},l,p}+d_{{\rm out},l,p}\bigr).
\]
Equal rank and adapter capacity across layers give the reduction factor $L/k$. Target sets differ between the Q/K/V mechanism study and the Q/K/V or Q/V bridge configurations in Appendix~\ref{app:base_adapt_agreement}.}

\subsection{Training Details}

\paragraph{Random Seeds.} \chg{08}{The original main results in \cref{tab:full_comparison,tab:generalization} use seed 42} for PyTorch, NumPy, and Python's random module. CUDA deterministic mode is enabled via \texttt{torch.backends.cudnn.deterministic = True}. \chg{08}{The supplementary evaluations in Appendix~\ref{app:multiseed} list their seed or run labels explicitly.}

\paragraph{Optimizer.} AdamW optimizer with learning rate $1\times10^{-4}$, weight decay 0.01, $\beta_1 = 0.9$, $\beta_2 = 0.999$, $\epsilon = 1\times10^{-8}$. No learning rate warmup is used. Training runs for 3 epochs on all configurations.

\paragraph{Batch Size.} Per-device batch size of 8 with gradient accumulation over 8 steps, yielding an effective batch size of 64 per training step (256 across 4 GPUs with DDP).

\paragraph{Precision.} FP16 mixed precision via PyTorch AMP. LoRA adapters are initialized in FP32 and cast to FP16 during forward/backward passes. The base model weights remain frozen in FP16 throughout.

\paragraph{Data Preprocessing.} All training samples are tokenized with a maximum sequence length of 512 tokens. Samples exceeding this length are truncated; shorter samples are padded to the nearest multiple of 8 for efficiency. Commonsense tasks use a multiple-choice format with answer extraction; math tasks use chain-of-thought prompting with answer parsing via regex.

\subsection{Evaluation Protocol}

\chg{14audit}{Evaluation uses no in-context examples. Commonsense results use answer extraction and task-level accuracy. Mathematical reasoning uses chain-of-thought prompting followed by task-specific answer parsing; GSM8K and SVAMP require numerical answers, whereas AQUA is multiple choice. Decoding is greedy (temperature$=0$).}

\subsection{Statistical Methodology}

\chg{08}{\paragraph{Across-Run Summaries.} In \cref{tab:multiseed}, for three displayed observations $x_1,x_2,x_3$, we report $\bar{x}=\frac{1}{3}\sum_i x_i$ and sample SD $s=\sqrt{\frac{1}{2}\sum_i(x_i-\bar{x})^2}$. Panel (a) summarizes paired accuracy differences; panel (b) summarizes each method's accuracy. These across-run SDs are distinct from bootstrap intervals over evaluation examples and do not by themselves imply statistical significance.}

\section{Additional Visualizations}
\label{app:visualizations}

\subsection{Q/K/V Projection Analysis}

\begin{figure}[ht]
    \centering
    \includegraphics[width=0.75\textwidth]{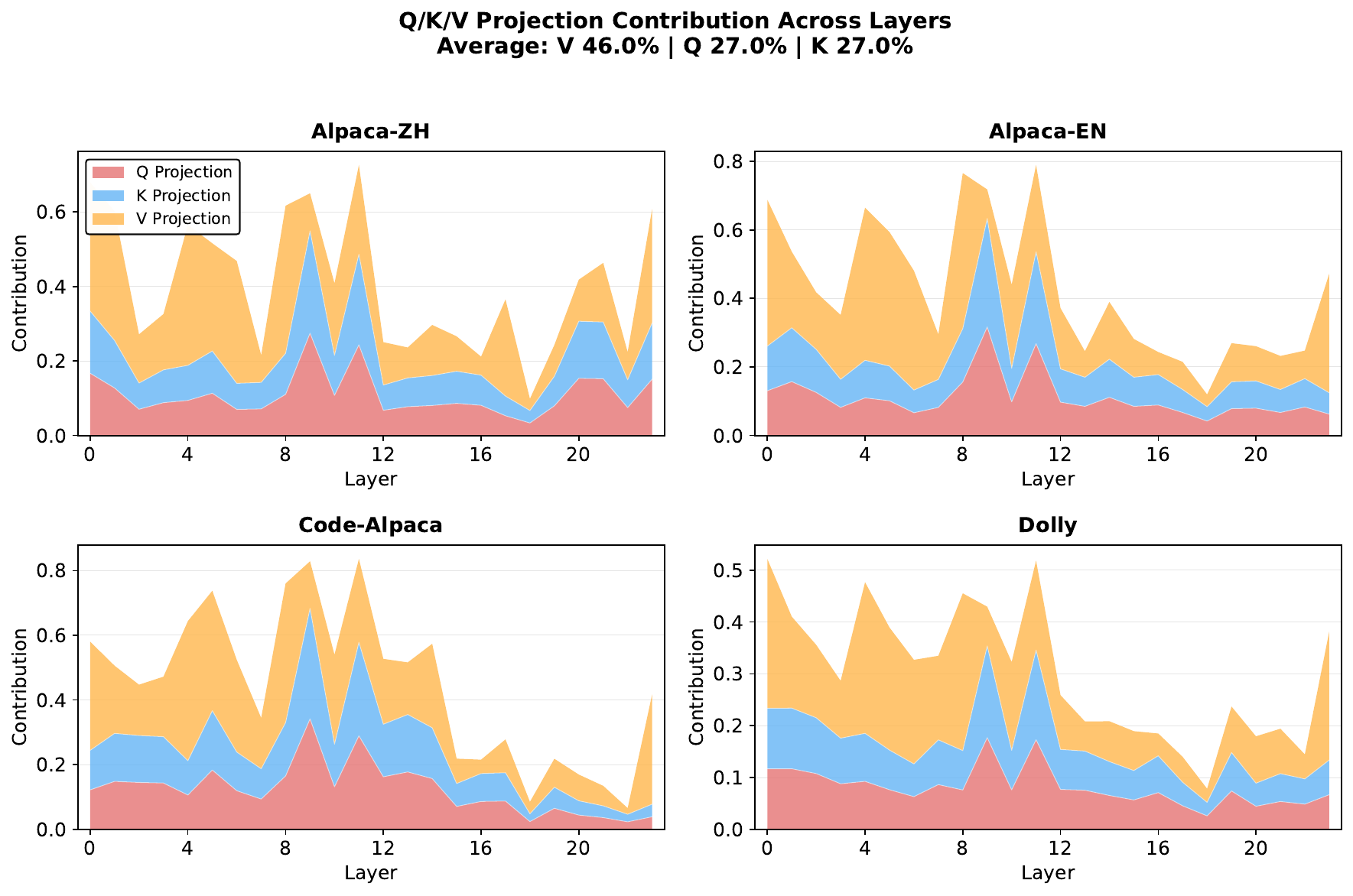}
    \caption{Contribution of Q, K, V projections within attention across four datasets. Value (V) dominates at 46\%, while Q and K contribute $\sim$27\% each.}
    \label{fig:qkv_dominance}
\end{figure}

\cref{fig:qkv_dominance} provides a detailed breakdown of how Q, K, and V projection adapters contribute to attention-mediated adaptation. The V projection consistently dominates across all four datasets, contributing 44.3\%--47.7\% of the total attention effect. \chg{14audit}{The common ordering across these four datasets complements their different layer-selection patterns.}

\subsection{Task-Specific Layer Selection}

\begin{figure}[ht]
    \centering
    \includegraphics[width=0.95\textwidth]{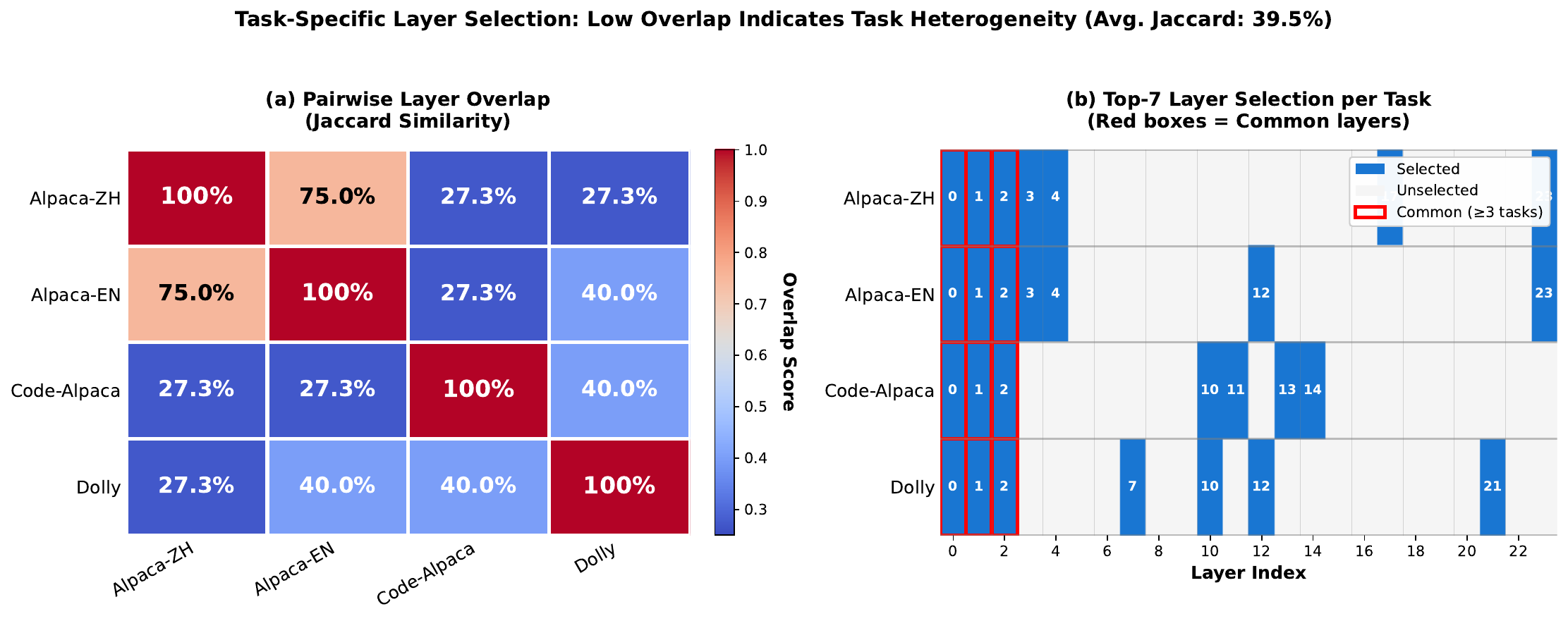}
    \caption{(a) Jaccard similarity matrix of top-7 layers across tasks (avg.\ 39.5\%). (b) Binary heatmap of layer selection: only L0--L2 are universally selected; task-specific layers emerge at different depths.}
    \label{fig:jaccard_detailed}
\end{figure}

\cref{fig:jaccard_detailed} visualizes the overlap between task-specific layer selections. \chg{13c}{Panel~(a) shows the pairwise Jaccard matrix: the highest overlap is 75.0\% between Alpaca-ZH and Alpaca-EN. The minimum, 27.3\%, occurs for Alpaca-ZH/Code-Alpaca, Alpaca-ZH/Dolly, and Alpaca-EN/Code-Alpaca. The average over all six task pairs is 39.5\%.} Panel~(b) provides a binary heatmap of which layers are selected for each task, making the task-specific divergence visually apparent.

\section{Calibration Analysis}
\label{app:calibration}

\subsection{Calibration Sample Sensitivity}

We investigate the sensitivity of \chg{01}{our frozen-base calibration} to \chg{01}{the number of calibration inputs $N$}. Using Qwen2.5-1.5B on the commonsense split, we vary $N \in \{10, 20, 30, 50, 100\}$ and measure both the resulting layer selection and downstream task performance at r16.

\begin{table}[ht]
\centering
\setlength{\tabcolsep}{6pt}
\renewcommand{\arraystretch}{1.05}
\caption{Effect of calibration sample size $N$ on calibration time and downstream accuracy (Qwen2.5-1.5B, r16, commonsense average).}
\label{tab:calibration_sensitivity}
\begin{changeblock}{14audit}
\begin{tabular}{@{}cccc@{}}
\toprule
\textbf{$N$} & \textbf{Time (s)} & \textbf{CS Avg.} & \textbf{$\Delta$ vs.\ $N{=}30$} \\
\midrule
10  & 14 & 59.4 & $-$0.4 \\
20  & 28 & 59.6 & $-$0.2 \\
30  & 42 & 59.8 & 0.0 \\
50  & 69 & 59.9 & +0.1 \\
100 & 137 & 59.7 & $-$0.1 \\
\bottomrule
\end{tabular}
\end{changeblock}
\end{table}

\chg{14audit}{Downstream accuracy spans 0.5 percentage points across this sample-size sweep, supporting $N=30$ as a useful cost--accuracy balance. Selection consistency across independent calibration draws is reported separately in \cref{tab:calibration_draws}.}

\subsection{Calibration Cost Breakdown}

The calibration overhead consists of three components: (1) sampling $N$ calibration examples from the training set (negligible), (2) \chg{01}{running $O(NL)$ gradient-free forward evaluations with independent component zero-ablation} to compute layer importance scores ($\sim$40 seconds for $N{=}30$ on Qwen2.5-1.5B with a single A100), and (3) sorting and selecting top-$k$ layers ($<$1 ms). \chg{05}{The approximately 40--42\,s calibration cost is less than 1\% of the 1.5--2 hours of fine-tuning in this configuration. \cref{tab:calibration_cost_stability} separately reports calibration alongside a shorter, 1,000-example training run.} Importantly, the calibration requires no gradient computation and can be performed in FP16 with minimal memory footprint.

\begin{table}[htbp]
\centering
\small
\renewcommand{\arraystretch}{1.05}
\caption{\chg{05}{Calibration cost and stability. Selection Jaccard is evaluated across five calibration draws with $N=30$ inputs. Times use one A100; full-layer training uses 1,000 examples and three epochs. These are the layer budgets used in this study.}}
\label{tab:calibration_cost_stability}
\begin{changeblock}{05}
\begin{tabular}{@{}lrrrr@{}}
\toprule
\textbf{Model} & $k$ & \textbf{Jaccard} & \textbf{Calibration (s)} & \textbf{Full LoRA (s)} \\
\midrule
Qwen-3B & 10 & 0.653 & 45 & 713 \\
Qwen-7B & 16 & 0.682 & 62 & 1,240 \\
LLaMA-3B & 8 & 0.747 & 29 & 714 \\
LLaMA-7B & 16 & 0.738 & 58 & 1,180 \\
\bottomrule
\end{tabular}
\end{changeblock}
\end{table}

\subsection{\chg{05}{Stability across Calibration Draws}}
\label{app:calibration_draws}

\chg{05}{\cref{tab:calibration_draws} repeats calibration with five independent draws. At $N=30$, selection Jaccard ranges from 0.617 to 0.747 across six models. The Qwen-3B sweep gains 0.129 from $N=10$ to $N=30$, versus 0.032 from $N=30$ to $N=100$. This provides additional evidence for a small calibration set. The layer budget $k$ is shown for each model; these experiments are separate from the fixed half-layer-budget study in Appendix~\ref{app:default_budget}. Unlike \cref{tab:calibration_sensitivity}, which compares each selection with an $N=30$ reference, this study measures stability across repeated calibration draws.}

\begin{table}[htbp]
\centering
\small
\renewcommand{\arraystretch}{1.05}
\caption{\chg{05}{Selection stability across five independent calibration draws. The first group compares six models at $N=30$; the second varies $N$ for Qwen-3B at $k=10$. The shared $N=30$ row is repeated for comparison.}}
\label{tab:calibration_draws}
\begin{changeblock}{05}
\begin{tabular}{@{}lrrr@{}}
\toprule
\textbf{Model} & $N$ & $k$ & \textbf{Jaccard} \\
\midrule
Qwen-0.5B & 30 & 7 & 0.653 \\
Qwen-1.5B & 30 & 8 & 0.617 \\
Qwen-3B & 30 & 10 & 0.653 \\
Qwen-7B & 30 & 16 & 0.682 \\
LLaMA-3B & 30 & 8 & 0.747 \\
LLaMA-7B & 30 & 16 & 0.738 \\
\midrule
Qwen-3B & 10 & 10 & 0.524 \\
Qwen-3B & 30 & 10 & 0.653 \\
Qwen-3B & 50 & 10 & 0.671 \\
Qwen-3B & 100 & 10 & 0.685 \\
\bottomrule
\end{tabular}
\end{changeblock}
\end{table}

\subsection{\chg{06}{Reference-Token Sensitivity}}
\label{app:token_sensitivity}

\chg{06}{Calibration uses the frozen base model's predicted token $t^*$ in \cref{eq:importance_aggregated}, so it requires no answer labels. On 500 labeled items, replacing this token with the gold-answer token yields a top-$k$ selection Jaccard of 0.81. The two reference choices yield downstream accuracies of 58.7\% and 58.5\%, differing by 0.2 percentage points. This check supports the predicted-token reference used by the method.}

\subsection{\chg{10}{Cross-Distribution Calibration}}
\label{app:cross_distribution}

\chg{10}{We vary the calibration source and training distribution for Qwen-3B at rank 16 with 5,000 training samples (\cref{tab:cross_distribution}). Across the three cross-distribution settings, layer-selection Jaccard relative to matched calibration is 0.453--0.512, and reported accuracy is 72.8--73.5\%. These results extend the calibration analysis beyond repeated draws from one distribution and show that useful layer selections transfer across the tested instruction distributions.}

\begin{table}[htbp]
\centering
\small
\renewcommand{\arraystretch}{1.05}
\caption{\chg{10}{Cross-distribution calibration on Qwen-3B ($r=16$, 5K training samples). Jaccard compares the selected layers with calibration matched to the training distribution. The final row gives the reported random-selection reference.}}
\label{tab:cross_distribution}
\begin{changeblock}{10}
\begin{tabular}{@{}llrr@{}}
\toprule
\textbf{Calibration source} & \textbf{Training target} & \textbf{Jaccard} & \textbf{Accuracy (\%)} \\
\midrule
Alpaca-EN & Alpaca-EN (matched) & 1.000 & 75.2 \\
Alpaca-EN & Code-Alpaca & 0.453 & 72.8 \\
Alpaca-EN & Dolly & 0.512 & 73.1 \\
Code-Alpaca & Alpaca-EN & 0.482 & 73.5 \\
\midrule
Random selection & --- & --- & 70.2 \\
\bottomrule
\end{tabular}
\end{changeblock}
\end{table}

\begin{changeblock}{03}
\subsection{Depth-Constrained Selection and Training Time}
\label{app:depth_timing}
Both operating modes rank layers by the frozen-base score in \cref{eq:importance_aggregated}. Accuracy mode selects over all layers. Efficiency mode restricts the candidates to $l\geq d^*$, allowing the frozen prefix below the actual first selected layer $d=\min\mathcal L_{\rm sel}$ to be reused. An uncached forward still evaluates all backbone layers. Frozen layers above an adapted layer must propagate gradients; only the entirely frozen prefix needs no backward graph.

For fixed processed inputs and a deterministic, dropout-free prefix, the depth-$d$ states are cached during the first epoch and reused subsequently. The host-memory cache scales as $n_{\rm train}T d_{\rm model}$ elements and is streamed per batch; $n_{\rm train}$ denotes training examples, not the calibration size $N$. Removing unselected adapter branches also saves their own forward, backward, and optimizer costs.

The controlled study uses a single H200, PyTorch 2.10, PEFT 0.18, bf16, AdamW, batch size eight, and sequence length 512. There is no gradient checkpointing, gradient accumulation, or DDP. Step measurements include forward, backward, and optimizer computation and are medians of 15 steps after five warm-ups. Table~\ref{tab:depth_speed} separates same-batch steps without prefix reuse, steady epochs with reuse, and three-epoch training totals including first-epoch cache construction. These training timings do not include the separately measured A100 calibration stage.
\end{changeblock}

\begin{table}[ht]
\centering\small\setlength{\tabcolsep}{3.5pt}
\caption{\chg{03}{Training speed depends on selected-layer depth. Single H200; nominal $k\approx L/3$. Speedups are relative to full LoRA. Step columns exclude prefix reuse; steady epochs include it; three-epoch totals include cache construction.}}
\label{tab:depth_speed}
\begin{changeblock}{03}
\begin{tabular}{@{}lrrrrrr@{}}
\toprule
& & Full step & $d=0$ & \multicolumn{3}{c}{$d=L/2$} \\
\cmidrule(lr){5-7}
Model & $L$ & (ms) & Step & Step & Steady ep. & 3 ep. \\
\midrule
Qwen2.5-0.5B & 24 & 62.22 & 1.131 & 1.412 & 1.461 & 1.444 \\
Qwen2.5-3B-Instruct & 36 & 168.78 & 1.097 & 1.446 & 1.900 & 1.720 \\
Llama-3.2-3B & 28 & 178.36 & 1.105 & 1.448 & 1.914 & 1.729 \\
Llama-3.1-8B & 32 & 325.85 & 1.078 & 1.412 & 1.934 & 1.722 \\
\bottomrule
\end{tabular}
\end{changeblock}
\end{table}

\chg{03}{At $d=0$, speedups are $1.08$--$1.13\times$. At $d=L/2$, the 3B and 8B models reach $1.90$--$1.93\times$ in steady epochs and $1.72$--$1.73\times$ over three epochs. These are efficiency-mode operating points; the unconstrained accuracy results use their own selected sets.}

\begin{table}[ht]
\centering\small\setlength{\tabcolsep}{4pt}
\caption{\chg{03}{Speed--accuracy operating points on Qwen2.5-0.5B ($k=8$, $r=16$, three epochs, two seeds). $d^*$ is the minimum allowed depth and $d$ the actual first selected layer. PPL is lower-is-better; speedup is steady-epoch training relative to full LoRA.}}
\label{tab:depth_frontier}
\begin{changeblock}{03}
\begin{tabular}{@{}lrrrr@{}}
\toprule
Mode & $d$ & Code-Alpaca PPL & Alpaca-EN PPL & Speedup \\
\midrule
Accuracy (unconstrained) & 0 & 2.294 & 4.821 & $1.13\times$ \\
Efficiency, $d^*=L/4$ & 7 & 2.301 & 4.854 & $1.35\times$ \\
Efficiency, $d^*=L/2$ & 12 & 2.308 & 4.907 & $1.46\times$ \\
Deepest-$k$ reference & 16 & 2.343 & 5.031 & $1.61\times$ \\
\bottomrule
\end{tabular}
\end{changeblock}
\end{table}

\chg{03}{For $d^*=L/4=6$, the actual first selected layer is $d=7$. The two PPL increases relative to accuracy mode are 0.3\% and 0.7\%. At $d^*=L/2$, the selected set is $\{12,13,14,15,20,21,22,23\}$ and the increases are 0.6\% and 1.8\%. The deepest-eight reference has larger increases (2.2\%, 4.4\%) and higher speed; its actual depth is different. These operating points show the quality--speed choice enabled by the common calibration ranking.}

\section{Cross-Architecture Comparison}
\label{app:cross_architecture}

\subsection{Layer Importance Patterns Across Families}

We compare layer importance patterns across three architecture families: Qwen2.5 (0.5B, 1.5B, 3B), LLaMA-3.2 (3B), and LLaMA-2 (7B). Despite differences in depth, width, and training data, several consistent patterns emerge:

\begin{enumerate}[leftmargin=1.5em, itemsep=2pt]
    \item \textbf{Early layer dominance.} In all architectures, the first 10--15\% of layers consistently rank in the top-5 by importance score, supporting Finding~3. This includes the embedding projection and early attention layers that establish token representations.
    \item \textbf{Middle layer task specialization.} Layers at 30--60\% depth show the highest variance in importance across tasks, consistent with Finding~4. These layers encode task-specific reasoning patterns that vary between commonsense and mathematical domains.
    \item \textbf{Depth-scaled selection.} \chg{14audit}{The main comparison in \cref{tab:generalization} uses 7/24, 8/28, 11/36, and 8/28 layers for the 0.5B, 1.5B, Qwen2.5-3B, and LLaMA 3B models, respectively. These are approximately 30\% layer budgets. The separate study below tests a fixed 50\% budget.}
\end{enumerate}

\subsection{Rank Scaling Behavior}

\chg{13b}{On Qwen2.5-1.5B, \cref{tab:full_comparison} compares four ranks and yields commonsense gains of 1.3, 2.0, 2.2, and 1.8 percentage points over full-layer LoRA. Selective adaptation therefore improves accuracy throughout the tested rank range, with the largest gain at r16. The cross-model comparison in \cref{tab:generalization} evaluates r16. Together, these experiments establish complementary evidence across ranks and model scales while retaining the parameter savings of layer selection.}

\subsection{MLP vs.\ Attention Contribution by Depth}

An important question for cross-architecture generalization is whether the MLP-dominance pattern (Finding 1) holds uniformly across layer depth or varies systematically. We decompose per-layer importance into MLP and attention components for Qwen2.5-1.5B and LLaMA-3-3B.

In both architectures, early layers (first 20\%) show the highest MLP/Attention ratio ($\sim$65/35), consistent with the hypothesis that early layers primarily modify feed-forward features. Middle layers (20--60\% depth) exhibit a more balanced ratio ($\sim$55/45), reflecting the co-importance of both information routing (attention) and feature transformation (MLP) for task-specific processing. Late layers (last 40\%) return to MLP dominance ($\sim$60/40), though with lower absolute importance scores.

This depth-dependent component balance has practical implications for future work on sub-layer selection: a hybrid strategy that selects MLP adapters broadly but attention adapters only in middle layers could achieve even higher efficiency than full-layer selection. \chg{14audit}{This motivates testing finer-grained allocation across components in future work.}

\subsection{\chg{07}{Default Layer Budget}}
\label{app:default_budget}

\chg{07}{\cref{tab:default_half_budget} selects the top half of layers by calibration score. This fixed fraction improves accuracy over full-layer LoRA on all eight models by 0.2--1.9 percentage points, without per-model layer-budget tuning.}

\begin{table}[htbp]
    \centering
    \renewcommand{\arraystretch}{1.05}
    \small
    \caption{\chg{07}{Fixed 50\% layer budget ($k=L/2$): rank $r=16$, 5,000 training examples. Accuracy (\%); $\Delta$: Adaptive-LoRA minus full-layer LoRA (percentage points).}}
    \label{tab:default_half_budget}
    \begin{changeblock}{07}
    \begin{tabular}{@{}lrrrcr@{}}
    \toprule
    \textbf{Model} & $L$ & $k$ & \textbf{Adaptive-LoRA} & \textbf{Full LoRA} & $\Delta$ \\
    \midrule
    Qwen-0.5B & 24 & 12 & 52.3 & 51.8 & +0.5 \\
    Qwen-1.5B & 28 & 14 & 56.2 & 55.8 & +0.4 \\
    Qwen-3B & 36 & 18 & 59.1 & 57.2 & +1.9 \\
    Qwen-7B & 32 & 16 & 80.0 & 79.8 & +0.2 \\
    Qwen-32B & 64 & 32 & 85.8 & 84.5 & +1.3 \\
    LLaMA-3B & 28 & 14 & 57.5 & 56.0 & +1.5 \\
    LLaMA-7B & 32 & 16 & 69.8 & 69.0 & +0.8 \\
    LLaMA-13B & 40 & 20 & 72.5 & 71.8 & +0.7 \\
    \bottomrule
    \end{tabular}
    \end{changeblock}
\end{table}

\section{Broader Impact}
\label{app:impact}

\subsection{Positive Societal Implications}

This work contributes to improving the efficiency of large language model adaptation, which has several positive societal implications:

\begin{enumerate}[leftmargin=1.5em, itemsep=2pt]
    \item \textbf{Democratized access.} By reducing the parameter count for fine-tuning by 3.5$\times$, \ours{} lowers the computational barrier for researchers and practitioners with limited GPU resources. Organizations that previously could not afford \chg{14audit}{full-layer LoRA adaptation} on larger models may now achieve comparable performance with significantly reduced hardware requirements.
    \item \textbf{Environmental sustainability.} \chg{14audit}{Smaller adapters reduce parameter storage. Training-time benefits depend additionally on selected-layer depth and prefix reuse, as quantified in Appendix~\ref{app:depth_timing}. Energy consumption is a separate system-level measurement.}
    \item \textbf{Multi-adapter serving.} Smaller adapter footprints enable serving more concurrent task-specific adapters on the same hardware, improving resource utilization in multi-tenant cloud environments and enabling personalized model deployment on edge devices.
    \item \textbf{Interpretability and transparency.} The mechanistic analysis underlying \ours{} provides interpretable insights into how LoRA adaptation distributes across model components. This transparency can help practitioners understand what their fine-tuned models have learned and where the adaptation has occurred, contributing to more responsible AI deployment.
\end{enumerate}

\subsection{Potential Risks and Mitigations}

We do not foresee direct negative societal impacts from this work, as it improves efficiency of existing fine-tuning methods without introducing new model capabilities or training paradigms. However, we acknowledge two indirect considerations:

\begin{itemize}[leftmargin=1.5em, itemsep=2pt]
    \item \textbf{Lowered cost of misuse.} By making fine-tuning more efficient, our method could marginally lower the cost of adapting models for harmful purposes. \chg{14audit}{Task-specific deployment therefore calls for evaluating safety behavior after adaptation, alongside task accuracy.}
    \item \textbf{Layer selection bias.} Our calibration-based selection relies on the distribution of calibration data. If calibration data is biased or unrepresentative, the selected layers may not generalize well to underrepresented populations. We mitigate this by using diverse, publicly available training datasets and validating across multiple task families.
\end{itemize}

\newpage
\section*{NeurIPS Paper Checklist}

\begin{enumerate}

\item {\bf Claims}
    \item[] Question: Do the main claims made in the abstract and introduction accurately reflect the paper's contributions and scope?
    \item[] Answer: \answerYes{}
    \item[] Justification: The abstract and introduction clearly state four contributions (mechanistic analysis, task-dependency finding, \ours{} method, empirical validation), all supported by experimental results in Sections~\ref{sec:mechanistic}--\ref{sec:experiments}.

\item {\bf Limitations}
    \item[] Question: Does the paper discuss the limitations of the work performed by the authors?
    \item[] Answer: \answerYes{}
    \item[] Justification: Limitations are discussed in Appendix~\ref{app:limitations}, covering selection granularity, task coverage, and theoretical foundations.

\item {\bf Theory assumptions and proofs}
    \item[] Question: For each theoretical result, does the paper provide the full set of assumptions and a complete (and correct) proof?
    \item[] Answer: \answerNA{}
    \item[] Justification: This paper is primarily empirical. The mechanistic analysis uses causal interventions with clearly stated procedures (Eqs.~\ref{eq:contribution}--\ref{eq:importance}).

\item {\bf Experimental result reproducibility}
    \item[] Question: Does the paper fully disclose all the information needed to reproduce the main experimental results of the paper to the extent that it affects the main claims and/or conclusions of the paper (regardless of whether the code and data are provided or not)?
    \item[] Answer: \answerYes{}
    \item[] Justification: \chg{14audit}{Model choices, training hyperparameters, dataset descriptions, evaluation metrics, and seed settings are provided} in Section~\ref{sec:experiments} and Appendix~\ref{app:implementation}.

\item {\bf Open access to data and code}
    \item[] Question: Does the paper provide open access to the data and code, with sufficient instructions to faithfully reproduce the main experimental results, as described in supplemental material?
    \item[] Answer: \chg{14audit}{\answerNo{}}
    \item[] Justification: \chg{14audit}{Code release is planned under the MIT license; this version does not provide a public code-release link. Public dataset sources are cited in the paper.}

\item {\bf Experimental setting/details}
    \item[] Question: Does the paper specify all the training and test details (e.g., data splits, hyperparameters, how they were chosen, type of optimizer) necessary to understand the results?
    \item[] Answer: \answerYes{}
    \item[] Justification: Full training configuration including optimizer, learning rate, batch size, epochs, and hardware are specified in Section~\ref{sec:experiments} and Appendix~\ref{app:implementation}.

\item {\bf Experiment statistical significance}
    \item[] Question: Does the paper report error bars suitably and correctly defined or other appropriate information about the statistical significance of the experiments?
    \item[] Answer: \answerYes{}
    \item[] Justification: \chg{14audit}{Appendix~\ref{app:multiseed} reports three-run means and sample SDs, separately for paired gains and absolute scores. Appendix~\ref{app:mlp_isolation} defines the 5,000-resample paired-bootstrap intervals for MLP-share gaps. These summaries do not assert significance for every single-seed comparison.}

\item {\bf Experiments compute resources}
    \item[] Question: For each experiment, does the paper provide sufficient information on the computer resources (type of compute workers, memory, time of execution) needed to reproduce the experiments?
    \item[] Answer: \answerYes{}
    \item[] Justification: \chg{14audit}{The hardware setup is described in Appendix~\ref{app:implementation}. Calibration costs are reported in Appendix~\ref{app:calibration}; the controlled step/epoch timing study specifies one H200 and its measurement protocol in Appendix~\ref{app:depth_timing}.}
    
\item {\bf Code of ethics}
    \item[] Question: Does the research conducted in the paper conform, in every respect, with the NeurIPS Code of Ethics \url{https://neurips.cc/public/EthicsGuidelines}?
    \item[] Answer: \answerYes{}
    \item[] Justification: This work focuses on improving parameter efficiency for LLM fine-tuning and poses no ethical concerns. All experiments use publicly available models and datasets.

\item {\bf Broader impacts}
    \item[] Question: Does the paper discuss both potential positive societal impacts and negative societal impacts of the work performed?
    \item[] Answer: \answerYes{}
    \item[] Justification: Broader impact is discussed in Appendix~\ref{app:impact}. Positive impacts include reduced computational costs for LLM adaptation, enabling broader access. \chg{14audit}{Potential misuse from lower adaptation costs and calibration-data bias is discussed there.}
    
\item {\bf Safeguards}
    \item[] Question: Does the paper describe safeguards that have been put in place for responsible release of data or models that have a high risk for misuse (e.g., pre-trained language models, image generators, or scraped datasets)?
    \item[] Answer: \answerNA{}
    \item[] Justification: This paper does not release pre-trained models or scraped datasets. The method applies to existing open-source models.

\item {\bf Licenses for existing assets}
    \item[] Question: Are the creators or original owners of assets (e.g., code, data, models), used in the paper, properly credited and are the license and terms of use explicitly mentioned and properly respected?
    \item[] Answer: \answerYes{}
    \item[] Justification: All datasets and models are properly cited. Qwen2.5 is used under its open-source license; all evaluation benchmarks are publicly available with their respective licenses respected.

\item {\bf New assets}
    \item[] Question: Are new assets introduced in the paper well documented and is the documentation provided alongside the assets?
    \item[] Answer: \answerNA{}
    \item[] Justification: No new datasets or pre-trained models are released. Code will be documented upon release.

\item {\bf Crowdsourcing and research with human subjects}
    \item[] Question: For crowdsourcing experiments and research with human subjects, does the paper include the full text of instructions given to participants and screenshots, if applicable, as well as details about compensation (if any)? 
    \item[] Answer: \answerNA{}
    \item[] Justification: This paper does not involve crowdsourcing or human subjects.

\item {\bf Institutional review board (IRB) approvals or equivalent for research with human subjects}
    \item[] Question: Does the paper describe potential risks incurred by study participants, whether such risks were disclosed to the subjects, and whether Institutional Review Board (IRB) approvals (or an equivalent approval/review based on the requirements of your country or institution) were obtained?
    \item[] Answer: \answerNA{}
    \item[] Justification: No human subjects research was conducted.

\item {\bf Declaration of LLM usage}
    \item[] Question: Does the paper describe the usage of LLMs if it is an important, original, or non-standard component of the core methods in this research? Note that if the LLM is used only for writing, editing, or formatting purposes and does \emph{not} impact the core methodology, scientific rigor, or originality of the research, declaration is not required.
    \item[] Answer: \answerNA{}
    \item[] Justification: LLMs are used as the subject of study (Qwen2.5 for fine-tuning), not as a tool for methodology development. No LLMs were used as non-standard components in the core methods.

\end{enumerate}

\end{document}